\documentclass{article}
\pdfoutput=1

\usepackage{PRIMEarxiv}

\usepackage[utf8]{inputenc} % allow utf-8 input
\usepackage{algorithm}  
\usepackage{algpseudocode}  
\usepackage[T1]{fontenc}    % use 8-bit T1 fonts
\usepackage{hyperref}       % hyperlinks
\usepackage{url}            % simple URL typesetting
\usepackage{booktabs}       % professional-quality tables
\usepackage{amsfonts}       % blackboard math symbols
\usepackage{nicefrac}       % compact symbols for 1/2, etc.
\usepackage{microtype}      % microtypography
\usepackage{lipsum}
\usepackage{fancyhdr}       % header
\usepackage{graphicx}       % graphics
\usepackage{amsmath}
\usepackage{color}
\usepackage{changepage}
\graphicspath{{media/}}     % organize your images and other figures under media/ folder
\algnewcommand{\Async}[1]{\State \textbf{Async}}
\algnewcommand{\Await}[1]{\State \textbf{Await}}
\title{
Retrieval Augmented Learning: A Retrial-based Large Language Model Self-Supervised Learning and Autonomous Knowledge Generation
\thanks{Corresponding author.}
}

\author{
  $\textbf{Zongyuan Li}^{1}, \textbf{Pengfei Li}^{1}, \textbf{Runnan Qi}^{2}, \textbf{Yanan Ni}^{2}, \textbf{Lumin Jiang}^{2}, \textbf{Hui Wu}^{1}$ \\
  $\textbf{Xuebo Zhang}^{1}, \textbf{Kuihua Huang}^{2,*}, \textbf{Xian Guo}^{1,*}$ \\
  1 College of Artificial Intelligence, Nankai University\\
  2 Laboratory for Big Data and Decision, National University of Defense Technology\\
}

\begin{document}
\maketitle

\vspace{-0.5cm}
\begin{adjustwidth}{1.5cm}{1.5cm}
\begin{abstract}
\vspace{0.4cm}

% The lack of domain-specific pre-training data fundamentally limits LLM-based decision systems in specialized applications, while post-train the model in the scenarios requires significant computational resources.
% In this paper, a train-free, reward-free, LLM self-supervised learning method called Retrial-Augmented Learning (RAL) is proposed. 
% The lack of pre-training knowledge in specialized scenarios severely hinders the application of Large Language Models (LLMs) based decision-making methods, while post-train a model in the scenarios requires significant computational resources.
The lack of domain-specific data in the pre-training of Large Language Models (LLMs) severely limits LLM-based decision systems in specialized applications, while post-training a model in the scenarios requires significant computational resources.
In this paper, we present Retrial-Augmented Learning (RAL), a reward-free self-supervised learning framework for LLMs that operates without model training.
By developing Retrieval-Augmented Generation (RAG) into a module for organizing intermediate data, we realized a three-stage autonomous knowledge generation of proposing a hypothesis, validating the hypothesis, and generating the knowledge. 
The method is evaluated in the LLM-PySC2\cite{LLM-PySC2} environment, a representative decision-making platform that combines sufficient complexity with domain-specific knowledge requirements. 
Experiments demonstrate that the proposed method effectively reduces hallucination by generating and utilizing validated knowledge, and increases decision-making performance at an extremely low cost. Meanwhile, the approach exhibits potential in out-of-distribution(OOD) tasks, robustness, and transferability, making it a cost-friendly but effective solution for decision-making problems and autonomous knowledge generation.
%  in unseen scenarios

\end{abstract}
\end{adjustwidth}
\vspace{0.2cm}

% keywords can be removed
% \keywords{Large Language Model \and StarCraft II \and Decision-making environment}

\section{Introduction}

% 第一段：大模型决策的优势
% 第二段：大模型决策面对的幻觉问题、领域知识缺失问题和实践知识缺失问题
% 第三段：传统的大模型决策方法大多通过优化提示词、构建工作流、重思考或者反思实现策略优化
% 第四段：少数资源丰富的研究团队使用模型训练方法进行策略训练，但他们大多在小规模的语言模型上进行训练
% 第五段：针对上面这些问题，我们提出了xxx方法
% 第六段：我们的贡献可以总结如下：（1）构建了文本域大模型自监督学习框架（2）提供了一种将RAG作为状态匹配方法、将知识库作为策略空间的思路（3）在实验中对比了这种自监督学习和相对于反思方法的提升

% Although LLM-based decision-making methods suffer from problems such as model hallucination and lack of domain knowledge,

\color{black} 
% % exhibit notable potential -> demonstrate distinct advantages
% In contrast to conventional Markov Decision Process (MDP) approaches, LLMs demonstrate distinct advantages in decision-making problems. Their abilities in language comprehension, context learning, and multi-tasking make them suitable for large-scale system integration, especially those requiring interpretability and interactivity.

% However, despite these advantages, LLMs face an inevitable problem: a lack of domain knowledge.
% The absence of decision-making instructions makes it inevitable for pre-trained models to face problems such as hallucination while training a large model in each scenario consumes unacceptable computing resources.
% These limitations consequently compel LLMs to operate on flawed knowledge representations. How to autonomously and computationally obtain accurate knowledge remains an unsolved problem. 

% Currently, most LLM decision-making methods do not learn in the environment, they use techniques such as prompt-engineering(\color{red}...\color{black}), multi-agent workflow(\color{red}...\color{black}) or RAG with fixed dataset(\color{red}...\color{black}). 
% These methods heavily relies on manually designed prompts, and result in limited applications in scenarios with enough tolerance for bad actions (\color{red}...\color{black}), with direct feedback of what needs to be done(\color{red}...\color{black}), or environment constructed with simplified action spaces(\color{red}...\color{black}).

In contrast to conventional Markov Decision Process (MDP) approaches, LLMs demonstrate distinct advantages in decision-making problems. Their abilities in language comprehension, context learning, and multi-tasking make them suitable for large-scale system integration, especially those requiring interpretability and human-AI interactivity.

However, despite these advantages, LLMs face an inevitable problem: a lack of domain knowledge.
The absence of decision-making instructions makes it inevitable for pre-trained models to face problems such as hallucination while training a large model in each scenario consumes unacceptable computing resources.
These limitations consequently compel LLMs to operate on flawed knowledge representations. How to autonomously and computationally obtain accurate knowledge remains an unsolved problem. 

% Some researchers use LLM reflection \cite{xu2023exploring}-\cite{MCghost} to increase performance. 
% Usually, reflection achieves better results than non-learning methods but still suffers from hallucination in the reflecting stage. 
% For environments such as MineCraft\cite{fan2022minedojo}, reflection is enough for knowledge generation, since the concepts(such as wool, diamond, and zombies) in the game are life-oriented and easy to understand. However for more general decision-making scenarios, the concepts and decision logic may not be mastered by the pre-trained models, and directly reflecting with inaccurate knowledge may lead to serious mistakes.

Currently, most LLM decision-making systems use techniques such as prompt-engineering \cite{2023arXiv231211865M}-\cite{li2025hierarchical}
, multi-agent workflow
\cite{ChatDev}-\cite{xiao2024tradingagents}
, or knowledge from a fixed dataset
\cite{ma2025vlms}\cite{fan2022minedojo}
and some researchers use LLM reflection \cite{xu2023exploring}-\cite{MCghost} to optimize the policy. 
Usually, reflection achieves better results than non-learning methods but still suffers from hallucination in the reflecting stage. 
For environments such as MineCraft\cite{fan2022minedojo}, reflection is enough for knowledge generation, since the concepts(such as wool, diamond, and sword) are common concepts and easy to understand. However for more general decision-making scenarios, the concepts and decision logic may not be mastered by the pre-trained models, and directly reflecting with inaccurate knowledge may lead to serious mistakes.

% the concepts and decision logic are more professional

% which is easy to understand, big model reflection can have a very obvious effect, because I already have enough knowledge.
% Some reflection methods only optimize the policy of the last step or episode, acting as a patch of the current policy instead of generating reusable knowledge for similar situations. 

% Some researchers use context learning methods named LLM reflection (\color{red}...\color{black}) to increase performance. 
% Usually, reflection achieves better results than non-learning methods, but still suffers from hallucination in the reflect stage.
% At the same time, reflection only optimizes the policy of the last step or episode, acting as a patch of the current policy instead of generating reusable knowledge for similar situations.

\begin{figure}[t]
  \centering
  \includegraphics[width=0.9\textwidth]{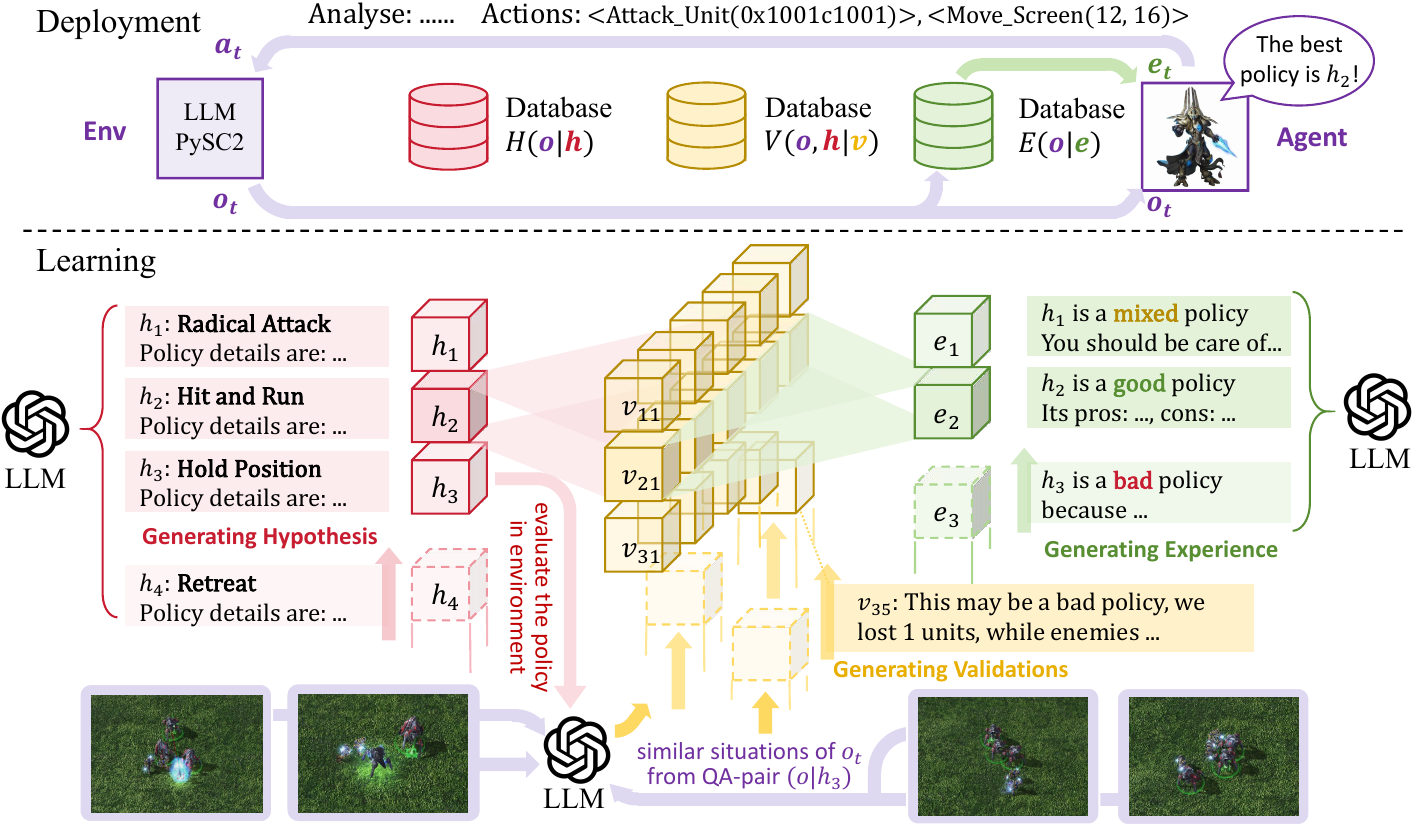}
  \caption{\textbf{General process of RAL.} 
  %%%An agent with LLM-SSCL framework has four modules: Action Module, Learning Module, Databases and Data Cache. 
  % At each step, the agent retrieve a list of hypothetical strategy or experience from Database $H(o|h)$ and $E(o|e)$, test the hypothetical policy or exploit the experience to make better decisions. At the same time, the agent learn from the state transition of last step, by proposing a different strategy, validate the current policy or present the fully validated knowledge into experience.
  In the learning process, the agent generates hypotheses about better policies, validates the hypothesis in similar situations, and summarizes the knowledge and experience when a policy has been sufficiently validated. When the proposed policies of a situation have been thoroughly evaluated, the agent directly uses the retrieved knowledge to make better decisions.
  }
  \label{fig1}
  \vspace{-0.7cm}
\end{figure}

% with enough computing resources  (more than 16x A100 or even 256x A100) 
At the same time, some researchers directly train models using supervised fine-tuning \cite{Diplomacy}\cite{MCLlamaRider} or reinforcement learning (RL) \cite{RLLLM}\cite{SMAC-R1}.
However, due to the extremely high demand for computing resources, the training process consumes a lot of GPU resources.
As a result, researchers who post-train a model must face a choice: either accept relatively low capabilities of smaller/quantilized/LoRA models or invest huge computing resources to complete the training of large models with hundreds of billions of parameters.

% these post-trained small models are unable to enjoy the benefits brought by the development of base model with hundreds of billions of parameters.

% In this paper, we develop RAG into a train-free, reward-free, self-supervised learning method called RAL, for knowledge generation in Markov decision process (MDP) that the LLM agent acts at each environment step according to its observation. 
% As illustrated in Fig.~\ref{fig1}, knowledge is generated in a three-stage process of \textbf{proposing a hypothesis}, \textbf{validating the hypothesis}, and \textbf{generating experience}, where RAG plays the role of extracting relevant knowledge and organizing intermediate data.
In this paper, we propose RAL, a train-free and reward-free learning framework that adapts RAG for organizing intermediate data, enabling LLM agents to dynamically generate knowledge through stepwise interactions with the environments.
As illustrated in Fig.~\ref{fig1}, the knowledge generation pipeline comprises three core operations: \textbf{hypothesis proposal}, \textbf{hypothesis validation}, and \textbf{knowledge generation}, with RAG functioning as both a retriever and intermediate data organizer.
It is worth noting that, the proposed learning framework is designed for remote models, making it quite cost-friendly, and compatible with edge devices, without the demand to deploy or train a model in hundreds of local GPUs.

% In RAL process, 
% We developed the RAG technique into a state matching and knowledge extraction module, store validated knowledge into databases, and reuse them in similar situations. 
% In our framework, knowledge is generated in a three-stage process of \textbf{proposing hypothesis}, \textbf{validating hypothesis}, and finally \textbf{generating experience}. 

In the experiments, we evaluated our method in the LLM-PySC2 environment and analyzed the performance during the training process. Results show that the proposed method effectively improves the decision-making ability of large models in an impressively short learning process. We also performed some out-of-distribution experiments and studied the robustness and data transferability of the method. 

% It is worth noting that parallel sampling is supported by our framework, which makes it possible to generate validated knowledge in a short time. Compared to LLM reflection and chain-of-thought (CoT), our method uses significantly fewer tokens in the deployment and is compatible with remote models, edge devices, and low-cost projects, without the demand to deploy or train a model in hundreds of local GPUs.

% It is worth noting that, the proposed learning framework is designed for remote models, making it quite cost-friendly and compatible with edge devices, without the demand to deploy or train a model in hundreds of GPUs.

Our contributions can be concluded as follows: 

\quad(1) We propose a self-supervised LLM learning framework for autonomous knowledge generation, providing a low-cost method that is train-free and reward-free for obtaining domain knowledge.

\quad(2) We develop RAG into a dynamic module to generate and organize intermediate learning data, breaking the traditional way of using fixed databases in the RAG process.

% \quad(3) We evaluate our method in the LLM-PySC2 platform, with experiments on OOD tasks, robustness, transferability of generated data, and the cost of prompt length and waiting time.
\quad(3) We evaluate our method in the LLM-PySC2 platform, with experiments on OOD tasks, robustness, transferability of generated data, and calculated statistics such as prompt length and waiting time.

In the last part of the article, we discussed the limitations and future works of our framework.
% We provide some directions for future works of improving the method.
More detailed materials like pseudo codes, prompts, and experiment settings can be found in the Appendices.
% View the Appendices for more detailed information if needed.

% Inspired by the idea of test-time scale(\color{red}...\color{black}) and retrial-based reasoning(\color{red}...\color{black}), we employed multiple validation in self-supervision stage, asking the agent to validate a hypothesis several times to reduce hallucination and possible noise of environment.
% Additionally, our method can be used as an auxiliary module for most advanced LLMs, which means, it is compatible with remote models, edge devices and low-cost projects, without the demand to deploy and train a model in hundreds of GPUs.

% In the last part of our article, we discussed the possible directions of improving the method. 
% Pseudo code, prompts and other details are list in appendix. View the appendix for more details if needed.

% with multiple verifications to reduce model hallucinations and realize relatively accurate self-supervision.
% Inspired by idea of test-time scale, we ask the agent to validate a hypothesis several times to relieve hallucination and possible noise.
% autonomously generate knowledge and restore into a database.

\begin{figure}[ht]
  \centering
  \includegraphics[width=0.9\textwidth]{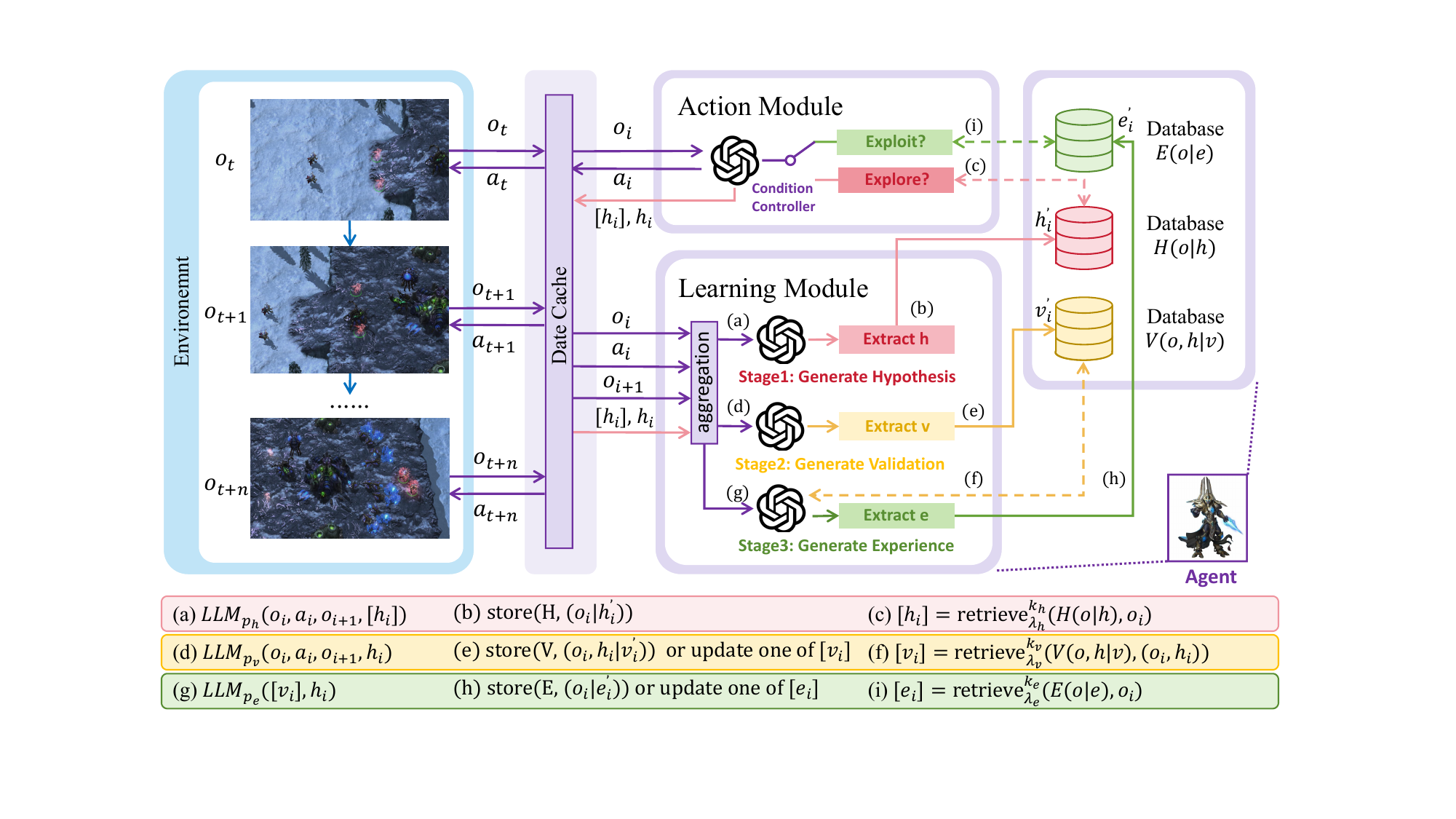}
  \caption{\textbf{RAL framework.} 
  %%%An agent with LLM-SSCL framework has four modules: Action Module, Learning Module, Databases and Data Cache. 
  At each step, the agent retrieve a list of hypothetical strategies or experience from Database $H(o|h)$ and $E(o|e)$, test the hypothetical policy or exploit the experience to make better decisions. At the same time, the agent learn from the state transition of last step, proposing a different strategy, validate the current strategy or present fully validated hypothetical strategy into experience. }
  \label{fig2}
\end{figure}

\section{Preliminaries}
\subsection{Markov Decision Process}

% 像传统的RL论文那样写即可，注意符号是文本形式时，给个特殊标记，然后加入一些LLM-pysc2的东西，如大模型决策的表示

Markov decision process refers to a process in which the state at any given moment is independent of future actions. In the past decade, it has been widely accepted as a fundamental setting of sequential decision-making problems such as optimum control, policy optimization, and multi-agent problems.

In a Markov decision process $MDP(S, A, R, \rho, s_0)$ initialized in $s_0$, the agent observes the environment state $s_t \in S$ and takes an action $a_t \in A$ at each step time $t$, where $S$ and $A$ refer to the state space and the action space. Then the environment returns a reward $r_{t}$ following the reward function $r_{t} = R(s_t, a_t)$, and transfer to a new state $s_{t+1}$ follows the generated possibility $p_{s_{t+1}}$ of the state transition function $p_{s_{t+1}} = \rho(s_t, a_t| s_{t+1})$. For partial observable problems, $MDP$ changes to $POMDP(O, S, A, R, \rho, s_0)$, where $O$ represents the observation space and $o_t \in O$ refers to the observed data at time $t$.

For LLM decision-making problems, LLM with profile $\widehat{p}$ receives a partial observable state $\widehat{o}_t$ and generates actions follows $\widehat{a}_t=LLM_p(\widehat{o}_t)$, where $\widehat{.}$ represents any data in text form. Considering that chain-of-thought (CoT) will be frequently used in the following parts, we define the CoT output as:
$$ (\widehat{c}_1,\widehat{c}_2,...\widehat{c}_n)_{CoT} = LLM_p(\widehat{.}_t) $$
where content $\widehat{c}_{i+1}$ is generated after $\widehat{c}_i$. 

\subsection{Retrieval Augmented Generation}

Retrieval Augmented Generation is a technique used to retrieve relevant text from a database and use the retrieved information to improve the quality of the generated content.
For query-answer(QA) pairs $(\widehat{q}|\widehat{a})$ and a QA database $D(\widehat{q}|\widehat{a})$, we define the retrieve action as:
$$[\widehat{a}] = retrieve_{\lambda}^k(D,\widehat{q})$$

% $$update(D,(\widehat{q}|\widehat{a}), \widehat{a}_0)$$  
In addition, we define the action for storing a QA pair as:
$$store(D,(\widehat{q}|\widehat{a}))$$
% and the function for obtaining the length of a list as:
% $$len([\widehat{a}])$$
Where $k$ refers to the top-k parameter that retains at most $k$ answers with the best matching scores, $\lambda$ refers to a threshold that only retains answers with scores higher than the threshold, and $[.]$ represents the retrieved list of answers. Accordingly, we use $len([.])$ to describe the length of $[.]$.

% 定义几个符号，如大模型生成、大模型生成思维链、RAG检索、RAG创建片段、RAG更新片段、数据库的符号，QA形式的符号

\section{Retrieval Augmented Learning}

% \subsection{Framework}

% In the proposed RAL framework, the agent completes knowledge generation in the \textbf{three stages}: generating hypothesis, generating validations, and presenting the results into knowledge or experience. as illustrated in Fig.~\ref{fig2}. 
In the RAL framework, knowledge is generated in the \textbf{three stages} process of \textbf{hypothesis proposal}, \textbf{hypothesis validation}, and \textbf{knowledge generation}. 
% As illustrated in Fig.~\ref{fig2}, we formalize the knowledge generation process in MDP scenarios as strategy optimization. Accordingly, formalize the three stages as "exploring a possible better strategy", "validating the strategy", and "presenting validations into an experience of a strategy". 
As illustrated in Fig.~\ref{fig2}, we formulate the learning process in MDP problems as a strategy optimization paradigm comprising: (1) strategy exploration (2) empirical validation, and (3) experience consolidation.
% we formalize the knowledge generation process in MDP scenarios as strategy optimization. formulate these stages as a strategy optimization paradigm comprising: 

In our framework, we define three databases $H(\widehat{s}|\widehat{h})$, $V((\widehat{s},\widehat{h})|\widehat{v})$ and $E(\widehat{s}|\widehat{e})$ for hypothesis, validation, and experience data, with top-k parameters $k_h$, $k_v$, $k_e$ and thresholds $\lambda_h$, $\lambda_v$, $\lambda_e$.  As shown in Fig~\ref{fig2}, at each step time, four LLMs work simultaneously in the RAL framework, among which one queries for actions, and the other three learning form interaction. Pseudo code of RAL can be found in Appendix A.

% \begin{figure}[ht]
%   \centering
%   \includegraphics[width=1.0\textwidth]{fig2.pdf}
%   \caption{\textbf{LLM-SSCL framework.} 
%   %%%An agent with LLM-SSCL framework has four modules: Action Module, Learning Module, Databases and Data Cache. 
%   At each step, the agent retrieve a list of hypothetical strategy or experience from Database $H(o|h)$ and $E(o|e)$, test the hypothetical policy or exploit the experience to make better decisions. At the same time, the agent learn from the state transition of last step, by proposing a different strategy, validate the current policy or present the fully validated knowledge into experience. }
%   \label{fig2}
% \end{figure}

% \begin{figure}[ht]
%   \centering
%   \includegraphics[width=0.9\textwidth]{fig-p-h.pdf}
%   \caption{\textbf{System Prompt for generating hypothetical policies.}}
%   \label{fig-prompt-h}
% \end{figure}

% \begin{figure}[ht]
%   \centering
%   \includegraphics[width=0.9\textwidth]{fig-p-v.pdf}
%   \caption{\textbf{System Prompt for generating validations of a proposed policy.}}
%   \label{fig-prompt-v}
% \end{figure}

% \begin{figure}[ht]
%   \centering
%   \includegraphics[width=0.9\textwidth]{fig-p-e.pdf}
%   \caption{\textbf{System Prompt for generating experience of a proposed policy.}}
%   \label{fig-prompt-e}
% \end{figure}

\subsection{Generating Hypothesis}

Exploring is necessary to obtain better strategies in policy optimization. As a basic part of self-supervised learning methods, it is usually performed by performing a random action \cite{DQN}-\cite{rashid2018qmixmonotonicvaluefunction} or adding Gaussian noise to the generated action\cite{ddpg}-\cite{contributors2021distar}.
However, these directionless exploration mechanisms suffer from the data efficiency that generates a large number of meaningless actions in the exploration process. 

\begin{figure}[ht]
  \centering
  \includegraphics[width=0.9\textwidth]{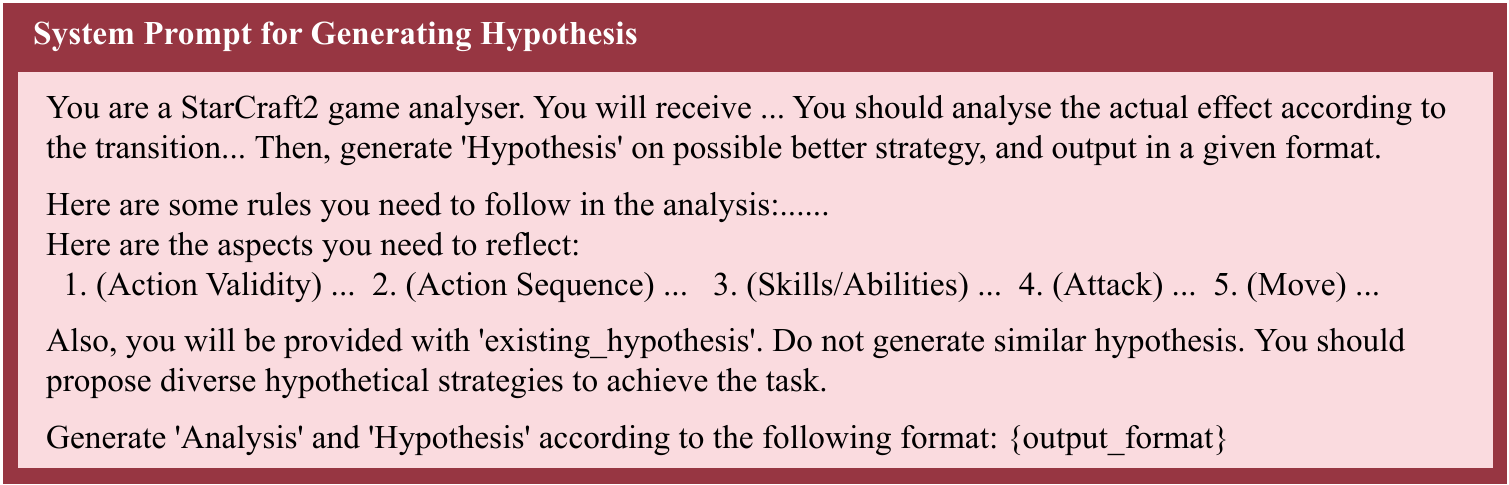}
  \caption{\textbf{System Prompt for generating hypothetical policies.}}
  \label{fig-prompt-h}
\end{figure}

Since it is capable for the language model to directly explore the policy space in certain directions, such noise-driven exploration mechanisms are unnecessary and also inconvenient. In our framework,  hypothesis proposal enables LLMs explore the policy space.
% For LLMs, such noise-driven exploration mechanisms are unnecessary and also inconvenient, Since it is capable of the language model to directly explore the policy space in certain directions. 
% We set the database $H(\widehat{o}|\widehat{h})$ as a hypothetical policy buffer. 
% To generate such observation-hypothesis pairs, a model analyzes the state transition $(\widehat{o}_t, \widehat{a}_t, \widehat{o}_{t+1})$ and proposes a better policy for its goal.
% The model is asked to generate a different policy from existing hypothetical policies $[\widehat{h}_t]$, to expand the explored policy space:
To generate observation-hypothesis pairs $(\widehat{o}|\widehat{h})$, a model analyzes the state transition $(\widehat{o}_t, \widehat{a}_t, \widehat{o}_{t+1})$ and proposes a better policy for its goal, and avoids generating a similar policy from existing hypothetical policies $[\widehat{h}_t]$:
% , to expand the explored policy space
$$(\widehat{c}_t^{analysis}, \widehat{h}_t)_{CoT} = LLM_{p_h}(\widehat{o}_t, \widehat{a}_t, \widehat{o}_{t+1}, [\widehat{h}_t])$$
and the generated pair $(\widehat{o}|\widehat{h})$ will be automatically store in the database for hypothetical policies $H(\widehat{o}|\widehat{h})$:
$$store(H,(\widehat{o}_t|\widehat{h}_t))$$
% This step is similar to traditional LLM reflection(\color{red}...\color{black}), however, the generated hypothetical policy will not be directly used in the deployment stage. These segments will only be retrieved for verification when the agent encounters similar situations during the learning process, which is used to reduce hallucinations and introduce knowledge through interactions with the environment.
This step shares similarities with traditional LLM reflection mechanisms, but differs in the usage of generated policies: rather than directly deploying the generated policies, the system only retrieves these policies for validity verification when encountering analogous scenarios during the learning phase.

% \begin{figure}[ht]
%   \centering
%   \includegraphics[width=0.9\textwidth]{fig-prompt-h.pdf}
%   \caption{\textbf{Prompt for generating hypothetical policies.}}
%   \label{fig-prompt-h}
% \end{figure}

% , due to the possible LLM hallucination

% This step is similar to traditional LLM reflection, but different from traditional LLM reflection(\color{red}...\color{black}), the proposed LLM exploration generates reusable knowledge, rather than a temporary patch of previous action.

\subsection{Generating Validation}

% The idea of LLM self-supervision inspired by the idea of traditional self-supervised learning and test-time scale(TTS) methods (\color{red}...\color{black}). 

% The key to self-supervised learning is how to introduce environmental feedback into the policy optimization process, instead of imitating ground-truth actions. 
% For RL methods, environment feedback is provided by the reward function, but for more general learning processes such as human learning, the policy optimization process is task-driven and reward-free.

% Validating a policy in the environment is the key to self-supervised learning. 
% Different from reward-driven mechanism, the agent assesses hypothetical policy by analyzing its impact on state transition and the consistency between state transition and the achievement of its goal. 
Validating a policy in the environment is the key to self-supervised learning. Unlike reward-driven mechanisms, the agent evaluates hypothetical policies by analyzing their influence on state transition dynamics and assessing the congruence between these transitions and goal achievement. This mechanism gives up explicit reward signals, making it a more general learning paradigm.
% Validating a policy in the environment is the key to learning from interactions. Just like calculating advantages in Actor-Critic based methods, the agent assess hypothetical policy on impact of the state transition and its goal, and a series of validations, make it more accurate for judging a strategy. This mechanism is driven by state transition and do not need explicit reward signals.
% The challenge in self-supervised learning lies in effectively integrating environmental feedback into policy optimization, rather than relying on imitation of ground-truth actions.
% While reinforcement learning (RL) relies on reward functions to formalize environmental feedback, more general learning paradigms operate through task-driven policy optimization without explicit reward signals.

\begin{figure}[ht]
  \centering
  \includegraphics[width=0.85\textwidth]{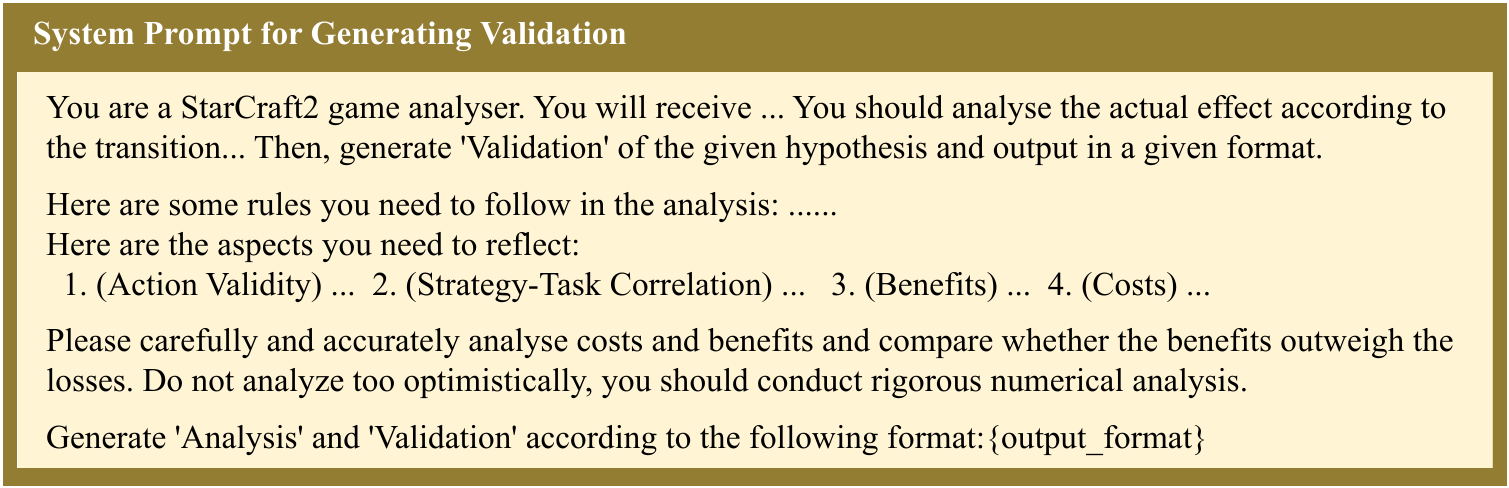}
  \caption{\textbf{System Prompt for generating validations of a proposed policy.}}
  \label{fig-prompt-v}
\end{figure}

% In the validation stage, the agent will try a hypothetical policy several times whenever encountered in similar situation when the policy proposed. 
% Through cycles of experimentation, the agent autonomously evaluates whether candidate strategies align with predefined objectives and quantitatively improve task performance, thereby enabling data-driven policy refinement.
% These data provides support for the following leanring stage, summarizing correct experiences and reducing model illusions in the future
% In the validation stage, the agent generates actions that follow the retrieved policy and assesses the influence of the policy at the next step. 

During validation, the agent executes policy-aligned actions and evaluates their causal impact on subsequent state transitions. Through a series of trials in similar situations, the agent collects comprehensive evaluations of the policies. These data support the next learning stage, which reduces hallucinations and provides accurate knowledge for the model.

At step time $t$, if there are not enough fine-prepared experiences and the agent retrieves a list of hypothetical policies $[\widehat{h}_t]$ from $H(\widehat{o}|\widehat{h})$ and generate actions follow one of the hypothetical policies: 
$$[\widehat{h}_t] = retrieve_{\lambda_h}^{k_h}(H,\widehat{o}_t)$$
$$(\widehat{c}_t^{analysis}, \widehat{a}_t^{explore})_{CoT} = LLM_{p_a}(\widehat{o}_t, random([\widehat{h}_t]))$$

we collect the transition data $(\widehat{o}_t, \widehat{a}_t^{explore}, \widehat{o}_{t+1})$, ask the LLM with profile $p_v$ for validation of the policy $\widehat{h}$ at step $t+1$, and then automatically store the $(\widehat{o},\widehat{h}|\widehat{v})$ pair in the RAG database:
$$(\widehat{c}_t^{analysis}, \widehat{v}_t)_{CoT} = LLM_{p_v}(\widehat{o}_t, \widehat{a}_t, \widehat{o}_{t+1}, \widehat{h}_t)$$
$$store(V,(\widehat{o}_t,\widehat{h}_t|\widehat{v}))$$

% Each validation data consists of several parts, such as general judgment, benefits, and costs. By separating the policy proposer and verifier, and requiring analysis from both cost-benefit perspectives during the verification process, it is possible to avoid prejudice of large models and obtain more accurate analysis.

% After generating the Q/A pair for policy validation, we collect the data and store into database $V(o,h|v)$:
% $$store(V,(\widehat{o}_t,\widehat{h}_t|\widehat{V}_t))\ with\ possibility\ \epsilon,\ $$
% $$else\ update(V,(\widehat{o}_t,\widehat{h}_t|\widehat{v}_t), random([\widehat{v}_t]))$$

% With the validations, the LLM can generate knowledge with much less hallucination at the final stage. Some imagined "good" policies will be proved to be "bad" in the validate process, according to the real impact of this policy. 

Through validation, our framework enables LLM to synthesize reliable knowledge with significantly reduced hallucinations in the final output. Some hypothetically optimal strategies, initially perceived as favorable, are empirically invalidated as suboptimal or even wrong policies when evaluated against their actual outcomes.

% \begin{figure}[ht]
%   \centering
%   \includegraphics[width=0.9\textwidth]{fig-prompt-v.pdf}
%   \caption{\textbf{Prompt for generating validations of a proposed policy.}}
%   \label{fig-prompt-v}
% \end{figure}

\subsection{Generating Experience / Knowledge}

Inspired by the idea of test-time scale\cite{SimpleTTS}-\cite{TTRL} and retrial-based reasoning\cite{RAGReasoning}, we enable an LLM to present all the collected data of a hypothetical policy into a piece of experience. The generated experience contains highly condensed information on the the costs, benefits, risks of the hypothetical policy, and details agent should pay attention to.
% , which will be used in similar situations in the deployment.
% As shown in Fig.~\ref{fig2}, 
The agent generates an experience for the retrieved hypothetical policy $\widehat{h}_t$ use its validations $[\widehat{v}_t]$:
% uses the hypothetical policy $\widehat{h}_t$ of the last step to retrieve its validations $[\widehat{v}_t]$, and generates an experience for the policy if there are enough validations:
$$[\widehat{v}_t] = retrieve_{\lambda_v}^{k_v}(V,(\widehat{o}_t, \widehat{h}_t))$$
$$(\widehat{c}_t^{analysis}, \widehat{e}_t)_{CoT} = LLM_{p_e}(\widehat{h}_t, [\widehat{v}_t])$$

% After generating the Q/A pair for policy experience, we collect the data and store into database $E(o|e)$:
% $$store(E,(\widehat{o}_t|\widehat{e}_t))\ with\ possibility\ \epsilon,\ $$
% $$else\ update(E,(\widehat{o}_t|\widehat{e}_t), \widehat{e}_t^{'})$$

\begin{figure}[th]
  \centering
  \includegraphics[width=0.85\textwidth]{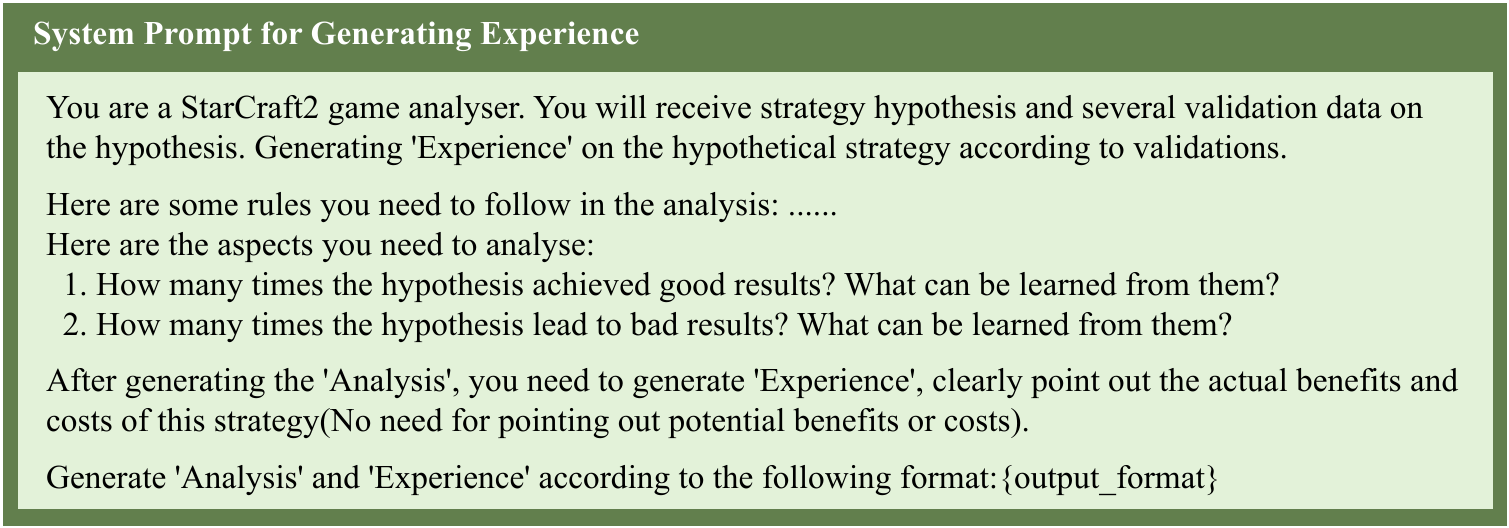}
  \caption{\textbf{System Prompt for generating experience of a proposed policy.}}
  \label{fig-prompt-e}
\end{figure}

As the learning process continues, the proposed hypothetical policies will be associated with the corresponding experiences. In a game step $t$, if enough policies have been validated and associated with experience $[\widehat{e}_t]$, the agent exploit all the retrieved experiences to deal with current situation:
$$[\widehat{e}_t] = retrieve_{\lambda_e}^{k_e}(E,\widehat{o}_t)$$
$$(\widehat{c}_t^{analysis}, \widehat{a}_t^{exploit})_{CoT} = LLM_{p_a}(\widehat{o}_t, [\widehat{e}_t])$$

When the agent makes decisions with the retrieved top-$k_e$ experiences, it actually uses $k_h \times k_v$ data, along with a lot of analysis in the CoT process. This mechanism significantly shortens the input prompt in the deployment, reduces the cost, storage, waiting time, and improves the data efficiency.

% \begin{figure}[ht]
%   \centering
%   \includegraphics[width=0.5\textwidth]{fig.pdf}
%   \caption{\textbf{LLM-PySC2 framwork.} In LLM-PySC2, the original PySC2 observation will transform into a text-form observation. LLM-generated text action can be recognized and transformed into PySC2 action functions, enabling LLMs to interact with the StarCraft II environment. }
%   \label{fig:framework}
% \end{figure}

\color{black}
\section{Experiments}

\color{black}

In experiments, we evaluate our method in the LLM-PySC2 environment, following the standard MDP settings. We studied the following questions:

\qquad \textbf{Q1}: Can the proposed method outperform traditional LLM reflection methods?

\qquad \textbf{Q2}: How much does decision-making ability improve in the learning process?

\qquad \textbf{Q3}: Are the generated experience data suitable for out-of-distribution (OOD) tasks?

\qquad \textbf{Q4}: Can the self-supervised learning method work with different LLMs? 

\qquad \textbf{Q5}: Can the data generated by one LLM be migrated to other models?

\qquad \textbf{Q6}: Does the method outperform other aspects, such as prompt length and waiting time?

% All the experiments are executed in LLM-PySC2 environment, due to its good support for micro operations and standard MDP settings. We conducted these experiments on Windows-11 systems, with RAG services provided by Dify and embedding service provided by GLM Embedding-3. All experiments were performed in the StarCraft II backend of version 9.0.14 (93333) and a developed version of LLM-PySC2 based on the released version v0.1. All the propmts are listed in Appendix B. More detailed experiment setting can be seen in Appendix C.

To answer questions Q1 (comparison with reflection), Q2 (improvement in the learning process), and Q3 (out-of-distribution performance), we conduct experiments on some of the LLM-PySC2 scenarios and LLM-SMAC that the text length is within the 8k limitation of GLM Embedding-3. For each scenario, we evaluate the decision-making ability at checkpoints of every 5 learning episodes, evaluating through 20 games with the learning module disabled.

To answer questions Q4 (robustness) and Q5 (data transferability), we conduct experiments on representative tasks, 2a\_harass from LLM-PySC2 and 3s\_vs\_3z from LLM-SMAC. With the same prompts, different models learn in the environment and are tested at each checkpoint of 5 episodes of learning. After 25 learning episodes, we provide other LLMs with the generated data and test their performance. 

Finally, we count statistical data for Q6, such as prompt length, waiting time, and resource occupation for different methods and models. Results show that our method outperforms direct decision-making, reflection, and long CoT reasoning with a negligible cost. All the prompts are listed in Appendix B. A more detailed experiment setting can be seen in Appendix C.

% Experimental results demonstrate that, even the proposed Naive RAL method lacks some functions such as utilizing long-horizon data, it still improves the decision ability in a significantly short learning process. Meanwhile, our method exhibits advantages in robustness, transferability, low disk occupancy, and low economic cost.

\subsection{Comparison with LLM Reflection (Q1)}

Given that reducing the hallucination of LLM reflection is one of the original motivations of our work, we tested our method and LLM reflection methods. We use GPT-3.5-turbo as the actor and learner, and correspondingly, we use text observation to adapt to the interfaces of the GPT model and RAG services. 

As shown in Table~\ref{tab1}, we tested direct LLM decision-making (\textbf{Baseline}), decision-making with last step reflection (\textbf{Reflection}), the performance of RAL after 25 episodes learning (\textbf{RAL-25}), the best performance in the learning process (\textbf{RAL-best}), and the improvement over baseline and reflection.
Three indicators are used to represent decision ability: Winning Rate($WR$), Value of Killed Units ($V_{killed}$) and Kill/Death ratio($KD$). These data are calculated as follows:

% $$ WR= 100\% \times n_{win} /n_{total} $$
$$ V_{unit} = minerals(unit) + 2 \times gas(unit)$$
$$ KD = V_{killed} / V_{lost} , WR= 100\% \times n_{win} /n_{total}$$

where $n_{win}$ and $n_{total}$ refer to the number of games won and in total. $minerals(unit)$ and $gas(unit)$ refer to the training resource of the unit. The gas resources are doubled during the unit value calculation, considering that the halved collection speed makes it twice the value of minerals.

% \begin{table}[ht]
% \vspace{0.2cm}
% \caption{Evaluation of Different context leanring method in LLM-PySC2 Tasks.}\label{tab1}
% \begin{center}
% \vspace{-0.2cm}
% \small
% \renewcommand\arraystretch{1.2}
% \begin{tabular}{p{3.7cm} p{1.5cm} p{1.5cm} p{1.5cm} p{1.7cm} p{1.7cm}}
% % \hline
% % & \multicolumn{6}{c}{Task name} \\
% \toprule
% Method / KD(WR) & 3s\_vs\_3z & 2a\_harass & 3ph\_harass & 4s\_vs\_1R4r & 4s\_vs\_5r\\
% \midrule
% Baseline           & 0.44 (35\%) & 0.90 (35\%) & 0.28 (5\%)  & 0.62 (5\%)  & 0.60 (30\%) \\
% Reflection         & 0.15 (0\%)  & 0.80 (40\%) & 0.22 (0\%)  & 0.63 (10\%) & 0.80 (\textbf{55\%}) \\
% % Reflection+RAG      & Any & Any & Any & Any & Any \\
% RAL-25(Ours)     & \textbf{1.17} (\textbf{95\%}) & 1.19 (70\%) & 0.39 (20\%) & \textbf{0.67} (\textbf{15\%}) & \textbf{0.81} (50\%) \\
% RAL-best(Ours)     & \textbf{1.17} (\textbf{95\%}) & \textbf{1.49} (\textbf{75\%}) & \textbf{0.42} (\textbf{25\%}) & \textbf{0.67} (\textbf{15\%}) & \textbf{0.81} (50\%) \\
% \midrule
% Improvement \\
% RAL-best over Baseline    & 0.73 (60\%) & 0.59 (40\%) & 0.14 (20\%) & 0.05 (10\%) & 0.21 (20\%) \\
% RAL-best over Reflection  & 1.02 (95\%) & 0.69 (35\%) & 0.20 (25\%) & 0.04 (5\%) & 0.01 (-5\%) \\

% \bottomrule
% \end{tabular}
% \end{center}
% \end{table}

\begin{table}[ht]
\caption{Evaluation of different context learning methods in LLM-PySC2 tasks} \label{tab1}
\centering
\small
\renewcommand\arraystretch{1.2}
\begin{tabular}{p{3.5cm} p{1.5cm} p{1.5cm} p{1.5cm} p{1.7cm} p{1.7cm}}
\toprule
    \multicolumn{5}{r}{Task Name / KD(WR) \qquad \qquad \qquad} \\
    \cmidrule(r){2-6}
Method & 3s\_vs\_3z & 2a\_harass & 3ph\_harass & 4s\_vs\_1R4r & 4s\_vs\_5r \\
\midrule
Baseline           & 0.44 (35\%) & 0.90 (35\%) & 0.28 (5\%)  & 0.62 (5\%)  & 0.60 (30\%) \\
Reflection         & 0.15 (0\%)  & 0.80 (40\%) & 0.22 (0\%)  & 0.63 (10\%) & 0.80 (\textbf{55\%}) \\
RAL-25(Ours)       & \textbf{1.17} (\textbf{95\%}) & 1.19 (70\%) & 0.39 (20\%) & \textbf{0.67} (\textbf{15\%}) & \textbf{0.81} (50\%) \\
RAL-best(Ours)     & \textbf{1.17} (\textbf{95\%}) & \textbf{1.49} (\textbf{75\%}) & \textbf{0.42} (\textbf{25\%}) & \textbf{0.67} (\textbf{15\%}) & \textbf{0.81} (50\%) \\
\midrule
Improvement \\
RAL-best over Baseline    & 0.73 (60\%) & 0.59 (40\%) & 0.14 (20\%) & 0.05 (10\%) & 0.21 (20\%) \\
RAL-best over Reflection  & 1.02 (95\%) & 0.69 (35\%) & 0.20 (25\%) & 0.04 (5\%) & 0.01 (-5\%) \\
\bottomrule
\end{tabular}
\vspace{-0.4cm}
\end{table}

Experimental evidence suggests that the absence of in-environment training limits LLMs' decision-making capabilities, resulting in suboptimal performance on assigned tasks. While reflection slightly improves LLM's decision-making ability in some scenarios, it performs significantly worse than our method. It is worth noting that LLM reflection even harms the performance in tasks such as 3s\_vs\_3z and 3ph\_harass due to its hallucination, preconception, or prejudice. It indicates that RAL reduces LLM hallucination and improves the decision ability in MDP problems.

\subsection{Improvement in Learning Process (Q2)}

In order to demonstrate the improvement more intuitively during the training process, we recorded the winning rate and value of killed units of 20 games after every 5 rounds of training. 

\begin{figure}[ht]
  \centering
  \includegraphics[width=1.0\textwidth]{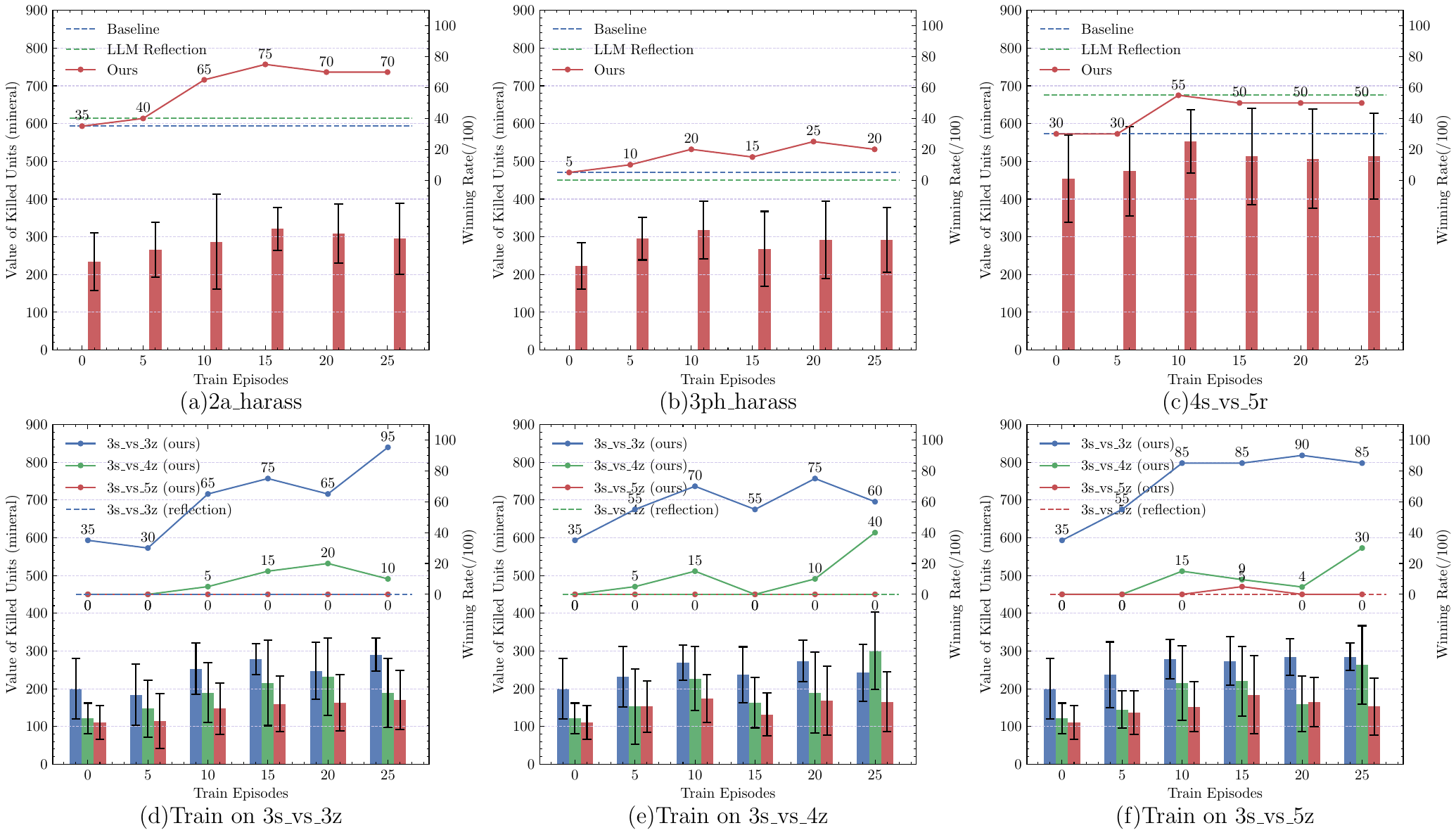}
  \caption{\textbf{Improvment of LLM's decision ability in the RAL learning process.} (a-b) Task of 2 Adepts harass Zerg base and 3 Pheonixes harass Zerg base, the goal of these harass tasks is to kill at least half of enemy workers; (c) Task of 4 Stalkers fight against 5 Zerg Roaches; (d-f) OOD tasks, train in one of the 3s\_vs\_nz tasks (3s\_vs\_3z, 3s\_vs\_4z, 3s\_vs\_5z), and evaluate performance in all the three scenarios.}
  \label{fig-data1}
\end{figure}

As shown in Figure.~\ref{fig-data1}, results demonstrate that our method increases decision ability in an extremely short process. Compared to the $10^{5}$ to $10^{7}$ step data for RL algorithms \cite{rashid2018qmixmonotonicvaluefunction}, our method saves at least 100x the interacting steps, thanks to LLMs' ability in semantic information processing, reasoning, and avoidance for obviously meaningless actions.

% We find that the more professional the scenario, the worse the effectiveness of direct decision-making and reflection. In LLM-PySC2 3ph\_harass task and all the three 3s\_vs\_nz tasks, reflection harms the performance. 

% \subsection{\textbf{Q3}: Are the learned experience data suitable for out-of-distribution tasks?}
\subsection{Out-of-distribution Performance (Q3)}

A problem with reinforcement learning algorithms is their weak OOD ability. When input vectors or task settings differ enough from those in training, it probably deteriorates the trained policy and harms the ability. However, with the ability to extract semantic information from natural language, LLMs handle different data well and perform well in OOD tasks.

As illustrated in (d) to (f) of Fig.~\ref{fig-data1}, RAL agent learns in a situation and is tested in similar tasks with different enemy units. Unlike RL methods, we find that RAL can learn in \textit{over difficulty} scenario 3s\_vs\_5z and performs better in similar easy task 3s\_vs\_3z, even unable to defeat the five units in the learning process. Also, we find that the generated experience of easy tasks can be used in similar tasks with more difficulty, e.g. learning in the 3s\_vs\_3z scenario improves the performance of 3s\_vs\_4z and 3s\_vs\_5z.

% \subsection{\textbf{Q4}: Does the self-supervised leanmethod work with different LLMs?}
\subsection{Robustness Across Models (Q4)}

To evaluate cross-model robustness, we conducted experiments on multiple models with the same settings and prompt. We tested three foundation models GPT-3.5-Turbo, GPT-4o-mini, DeepSeek-V3, and a reasoning model DeepSeek-R1, visualized the performance of learning in 2s\_harass and 3s\_vs\_3z.

\begin{figure}[ht]
  \centering
  \includegraphics[width=0.90\textwidth]{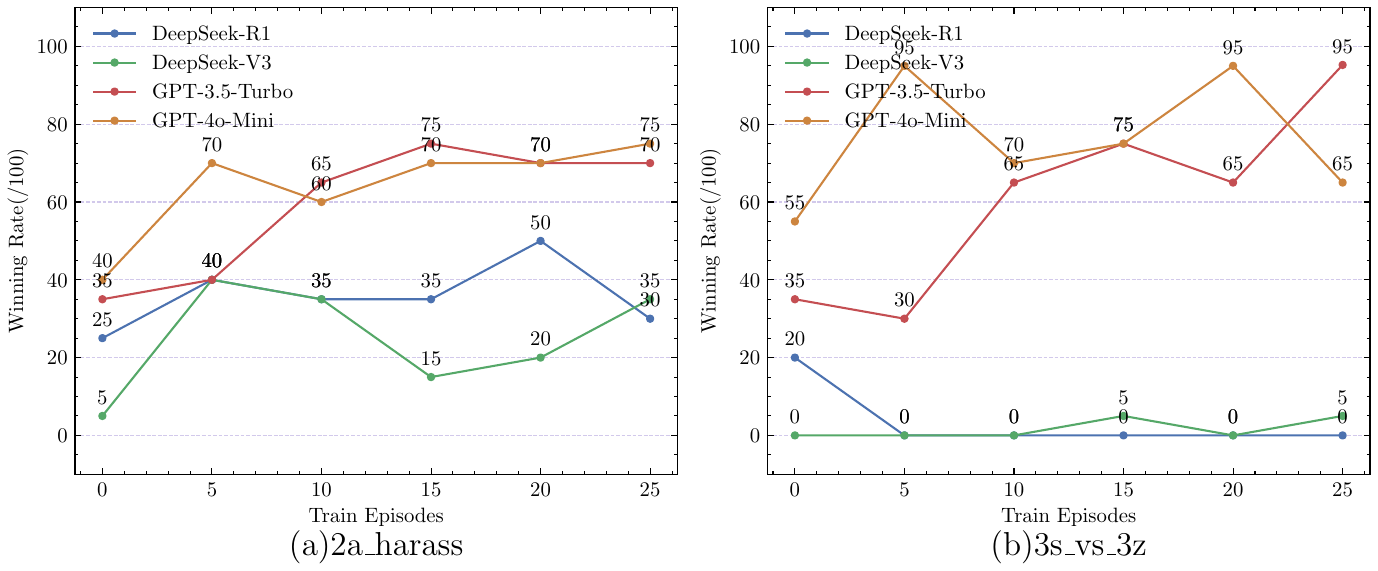}
  \caption{\textbf{RAL learning process for different models.} (a) Four models learn in the task of 2a\_harass (b) Four models learn in 3s\_vs\_3z. Generally speeking, it is possible for RAL to improve the strategic performance for different models, but the robustness is influenced by the model's ability and hallucination.}
  \label{fig-data2}
\end{figure}

As shown in Figure.~\ref{fig-data2}, all three foundation models achieve better results than direct decision-making (at episode 0), and the improvement scale depends on the models' original ability. Meanwhile, it is interesting that the reasoning model DeepSeek-R1 performs worse than GPT models, especially the GPT model with RAL-generated experience after the learning process. Reasoning models may focus more on their own experiences and face more hallucinations when they learn from the interactions.

\vspace{0.5cm}

% \subsection{\textbf{Q5}: Can the data generated by one LLM be migrated to other models?}
\subsection{Data Transferability (Q5)}

To further investigate the compatibility between models and the RAL framework, we tested the transferability of the RAL-generated experience. In the experiment, the four models use generated experience of 25 episodes of learning of other models and compare with the performance of direct decision making. 

As shown in Table.~\ref{tab2} and Table.~\ref{tab3}, here are some conclusions:
(1) GPT models work better in using the data generated by other models but achieve better results using the experience generated by themselves. 
(2) DeepSeek models are not good at using RAL-generated experience, even if the experience proved effective in improving performance. 
(3) Foundation model DeepSeek-V3 performs better than its reasoning model DeepSeek-R1 (with the same parameter size) after the learning process, indicating that the post-trained thinking mode may harm the RAL performance, hindering achieving better results in scenarios that require domain knowledge.

For the proposed methods, data transferability depends on the quality of the data and the model that uses the generated data. Well-generated data enables other models to directly improve their performance, and a good data user achieves better results using the same batch of experience.

\begin{table}[ht]
\caption{Transferability evaluation in 2a\_harass task with 25 episodes learning} \label{tab2}
\centering
\small
\renewcommand\arraystretch{1.2}
\begin{tabular}{p{3.5cm} p{2.1cm} p{2.1cm} p{2.1cm} p{2.0cm}}
\toprule
    \multicolumn{4}{r}{Model Name / KD(WR) \qquad \quad \quad} \\
    \cmidrule(r){2-5}
Learner Model Name  & GPT-3.5-Turbo & GPT-4o-Mini & Deepseek-V3 & Deepseek-R1 \\
\midrule
Without Experience           & 0.90 (35\%) & 0.98 (40\%) & 0.45 (5\%)  & 0.73 (25\%)  \\
GPT-3.5-Turbo (Learner)      & \textcolor{blue}{1.19 (70\%)} & 1.56 (75\%) & 0.82 (30\%) & 1.22 (65\%)  \\
GPT-4o-Mini (Learner)        & 1.31 (65\%) & \textcolor{blue}{1.33 (75\%)} & 0.59 (5\%)  & 0.91 (35\%)  \\
Deepseek-V3 (Learner)        & 1.27 (65\%) & 1.31 (75\%) & \textcolor{blue}{0.80 (35\%)} & 0.89 (30\%)  \\
Deepseek-R1 (Learner)        & 1.52 (60\%) & 1.09 (75\%) & 0.74 (30\%) & \textcolor{blue}{0.60 (30\%)}  \\
\bottomrule
\end{tabular}
\end{table}

\begin{table}[ht]
\caption{Transferability evaluation in 3s\_vs\_3z task with 25 episodes learning} \label{tab3}
\centering
\small
\renewcommand\arraystretch{1.2}
\begin{tabular}{p{3.5cm} p{2.1cm} p{2.1cm} p{2.1cm} p{2.0cm}}
\toprule
    \multicolumn{4}{r}{Model Name / KD(WR) \qquad \quad \quad} \\
    \cmidrule(r){2-5}
Model Name  & GPT-3.5-Turbo & GPT-4o-Mini & Deepseek-V3 & Deepseek-R1 \\
\midrule
Without Experience            & 0.44 (35\%) & 0.56 (55\%) & 0.12 (0\%)  & 0.28 (20\%)  \\
GPT-3.5-Turbo (Learner)       & \textcolor{blue}{1.17 (95\%)} & 0.66 (55\%) & 0.23 (10\%) & 0.20 (10\%)  \\
GPT-4o-Mini (Learner)         & 0.83 (75\%) & \textcolor{blue}{0.74 (65\%)} & 0.23 (15\%) & 0.11 (0\%)  \\
Deepseek-V3 (Learner)         & 0.41 (40\%) & 0.49 (45\%) & \textcolor{blue}{0.22 (5\%)} & 0.16 (0\%)  \\
Deepseek-R1 (Learner)         & 0.43 (40\%) & 0.36 (30\%) & 0.21 (5\%) & \textcolor{blue}{0.15 (0\%)}  \\
\bottomrule
\end{tabular}
\end{table}

% \subsection{Technical data statistics}

% \clearpage
\subsection{Advantages in Cost, Waiting Time and Resource Occupation (Q6)}

RAL demonstrates significant advantages in cost, specifically, prompt length, waiting time, and resource occupation. In the decision-making process, RAL introduces the retrieved experience into the input prompt and only increases the prompt length by about 300 tokens. At the same time, the cost of 25 episodes of learning is significantly less than post-trained the model, making it quite cost-friendly in solving MDP problems.

As shown in Table.~\ref{tab4}, input tokens, output tokens, and waiting time of RAL are nearly the same as direct decision making, while the reflection of the LLM and the direct use of a reasoning model cost much more. Compared to RAL, LLM reflection requires conducting additional interactions before making decisions, and reasoning models such as DeepSeek-R1 waste even more time and tokens.

\begin{table}[ht]
\vspace{0.2cm}
\caption{LLM Query Cost in 3s\_vs\_3z Decision Making Scenario}\label{tab4}
\begin{center}
\vspace{-0.2cm}
\small
\renewcommand\arraystretch{1.2}
\begin{tabular}{p{4.4cm} p{2.4cm} p{2.4cm} p{2.4cm} p{1.0cm}}
\toprule
Method / Value (Relative Value) & Input Tokens  & Output Tokens & Waiting Time & WR\\
\midrule
GPT3.5+Direct        & 4042.4 (1.00) & 867.6 (1.00) & 22.45s (1.00) & 35\%\\
GPT3.5+Reflection    & 7809.7 (1.93) & 2811.4 (3.24) & 76.39s (3.40) & 0\%\\
GPT3.5+RAL(Ours)     & 4391.4 (1.07) & 893.3 (1.02) & 24.04s (1.07) & 95\%\\
DeepSeek-V3 + Direct   & 4079.4 (1.01) & 685.5 (0.79) & 31.71s (1.41) & 0\%\\
DeepSeek-R1 + Direct   & 4019.4 (0.99) & 4655.9 (5.37) & 104.64s (4.66) & 20\%\\
\bottomrule
\end{tabular}
\end{center}
\end{table}

It is worth noting that, our method increases only several MBs of disk occupation for storing the generated text data. 
Without deploying a large model on the local devices, it is also possible to deploy the RAL framework on edge devices with remote LLMs, making it quite friendly for low-cost applications.

% \begin{table}[ht]
% \vspace{0.2cm}
% \caption{LLM Query Cost in Decision Making Process}\label{tab4}
% \begin{center}
% \vspace{-0.2cm}
% \small
% \renewcommand\arraystretch{1.2}
% \begin{tabular}{p{3.2cm} p{3.2cm} p{3.2cm} p{3.2cm}}
% \toprule
% Method  & Token Input (relative) & Token Output (relative) & Waiting Time (relative)\\
% \midrule
% GPT3.5+Direct        & 4042.4 (1.00) & 867.6 (1.00) & 22.45s \\ 
% \bottomrule
% \end{tabular}
% \end{center}
% \end{table}

\section{Discussion}

Although the proposed RAL method only learns in a one-step state transition, it is already possible to generate accurate knowledge in unseen environments. Strictly speaking, learning in a one-step state transition is not sufficient for generating optimal strategy that needs to be validated on long-horizon data. In this section, three methodological limitations will be discussed, followed by concluding remarks on potential future works.

% In this section, several limitations will be discussed, and conclusion will be made at last.

% Although the experimental results exhibit the potential in reducing hallucination, improving decision-making abilities, robustness across models, data transferability, and cost control, the current RAL framework is not well developed. Possible improvements include but are not limited to the following aspects: (1) utilize the environment reward; (2) reflect on long-horizon data; (3) policy iteration. We have not yet introduced these elements into the RAL framework, because of possible deterioration may also be introduced into the learning process along with these techniques.

\subsection{Limitations}

% The RAL method is limited by the performance of the model used in the learning process, the absence of environmental rewards, long-term data, and policy interaction. As a result, there is no guarantee that the framework can always find the correct policy. Meanwhile, the lack of long-term data and environmental reward makes it difficult to accurately determine the long-term impact of each policy.

% with not being able to explore the correct strategy, not being able to correctly judge the correctness of the strategy, and not being able to quantitatively judge the quality of the strategy.

\textbf{Insufficient exploration.} 

For decision-making problems, the size of the explored policy space limits the performance of the final policy. 
However, the LLM-learner may not be imaginative enough to find the optimal policy. 
The exploration ability of LLMs will directly affect the effectiveness of our framework. How to promote LLM exploration remains an unsolved problem.

\textbf{Insufficient validation for long-horizon policy.} 

The lack of long-term data, such as episode reward and multi-step state transition, limits the accuracy of evaluating the long-term performance of a policy. In the short term, some policies' costs outweigh benefits, but is crucial for the success of the task in the long term. However, long-term data will greatly increase the length of input prompt and introduce the confidence allocation problem into the learning process. The over-lengthy prompt may harm the performance, because LLMs have to analyze much more and may make more mistakes in the process.

\subsection{Conclusion}

In this paper, we propose RAL, an RAG-based framework for autonomously generating validated knowledge in MDP decision-making problems. 
We introduce the overall structure and key operations, 
The performance of the method is evaluated in the performance in LLM-PySC environment, a representative environment that is complex enough and do not engage the pre-training process of LLMs.
In the experiments, we found that RAL reduces more hallucinations, compared to direct LLM decision making and traditional LLM reflection, and performs well in OOD tasks. We also tested the robustness of different models and the transferability of the generated data. During the evaluation process, we accidentally discover that Deepseek model performs worse than old models such as GPT-3.5, indicating that CoT reasoning in incomplete or inaccurate knowledge will seriously deteriorate the decision-making and context learning ability of LLMs. In the last part, we discuss the limitations of RAL framework, provide some ideas for future works.

\appendix
\clearpage

\section*{Appendix A. Pseudo Code}
\setcounter{figure}{0}
\renewcommand{\thefigure}{A\arabic{figure}}

\subsection*{A.1 RAL Framework}
\begin{algorithm}
\caption{RAL: A Retrieval Augmented Self-Supervised Context Learning}
\begin{algorithmic}

\Require LLM decision-making prompt $\widehat{p}_a$. LLM learning prompt $\widehat{p}_h$, $\widehat{p}_v$, $\widehat{p}_e$. Q/A Database $H(\widehat{h}|\widehat{s})$, $V(\widehat{v}|(\widehat{s},\widehat{h}))$ and $E(\widehat{e}|\widehat{s})$ for hypothesis, validation and experience. Retrieve-controlling coefficient $k_h$, $k_v$, $k_e$ and $\lambda_h$, $\lambda_v$, $\lambda_e$. Update rate $\epsilon$. Parallel size $n$.

\State 
\State Initilalize environments Env$_1$ to Env$_n$
\For {$i$ in range($n$)}

\While{not Env$_i$.is\_terminated()}
    \State $\widehat{o_t}$ = Env$_i$.get\_obs()
    \State $[\widehat{e}]_t$ = $retrieve_{\lambda_e}^{k_e}(E, \widehat{o_t})$
    \State $[\widehat{h}]_t$ = $retrieve_{\lambda_h}^{k_h}(H, \widehat{o_t})$
    \If {t > 1 and $h_{t-1}$ exist}
        \State $[\widehat{v}]_t$ = $retrieve_{\lambda_v}^{k_v}(V, (\widehat{o}_{t-1}, \widehat{h}_{t-1}))$
    \EndIf
    
    \State 
    \State initialize action threads $p_a$
    \If {$len([\widehat{e}]_t) = k_e$}
        \State $p_a = $thread$(LLM_{p_a}(\widehat{o}_t, [\widehat{e}]_t))$
    \ElsIf {$len([\widehat{h}]_t) > 0$}
        \State $p_a = $thread$(LLM_{p_a}(\widehat{o}_t, \widehat{h}_t))$, where $\widehat{h}_t = random([\widehat{h}]_{t})$
    \Else
        \State $p_a = $thread$(LLM_{p_a}(\widehat{o}_t))$ 
    \EndIf
    
    \State 
    \State initialize learning threads $p_h$, $p_v$, $p_e$
    \If {$len([ht]) < k_h$}
        \State $p_h = $thread$(LLM_{p_h}(\widehat{o}_{t-1}, \widehat{a}_{t-1}, \widehat{o}_{t}, [\widehat{h}]_{t-1}))$
    \EndIf
    \If {$h_{t-1}$ exist and $len([\widehat{v}]_t) < k_v$}
        \State $p_v = $thread$(LLM_{p_v}(\widehat{o}_{t-1}, \widehat{a}_{t-1}, \widehat{o}_{t}, \widehat{h}_{t-1}))$
    \EndIf
    \If {$len([\widehat{e}]_t) < k_e$ and $len([\widehat{v}]_t)=k_v$}
        \State $p_e = $thread$(LLM_{p_e}([\widehat{h}]_{t-1}, [\widehat{v}]_t))$
    \EndIf 

    \State 
    \State start threads $p_a$, $p_h$, $p_v$, $p_e$ and waiting for response $\widehat{a}_t$, $\widehat{h}_{t-1}^{'}$, $\widehat{v}_{t-1}^{'}$, $\widehat{e}_{t-1}^{'}$
    \State create a new segment $store(H, (\widehat{o}_{t-1}| \widehat{h}_{t-1}^{'}))$ if response $\widehat{h}_{t-1}^{'}$ exist
    \If {random() < $\epsilon$}
        \State update a retrieved segment $update(V, (\widehat{o}_{t-1}, \widehat{h}_{t-1}| \widehat{v}_{t-1}^{'}))$ if response $\widehat{v}_{t-1}^{'}$ exist
        \State update a retrieved segment $update(E, (\widehat{o}_{t-1}| \widehat{e}_{t-1}^{'}))$ if response $\widehat{e}_{t-1}^{'}$ exist
    \Else
        \State create a new segment $store(V, (\widehat{o}_{t-1}, \widehat{h}_{t-1}| \widehat{v}_{t-1}^{'}))$ if response $\widehat{v}_{t-1}^{'}$ exist
        \State create a new segment $store(E, (\widehat{o}_{t-1}| \widehat{e}_{t-1}^{'}))$ if response $\widehat{e}_{t-1}^{'}$ exist
    \EndIf

    \State 
    \State Env$_i$.step($\widehat{a}_t$)
\EndWhile
\EndFor

\end{algorithmic}
\end{algorithm}

\clearpage
\section*{Appendix B. All Prompt}
\setcounter{figure}{0}
\renewcommand{\thefigure}{B\arabic{figure}}

\subsection*{B.1 example of observation}

% \vspace{-0.2cm}
\begin{figure}[ht]
  \centering
  \includegraphics[width=0.9\textwidth]{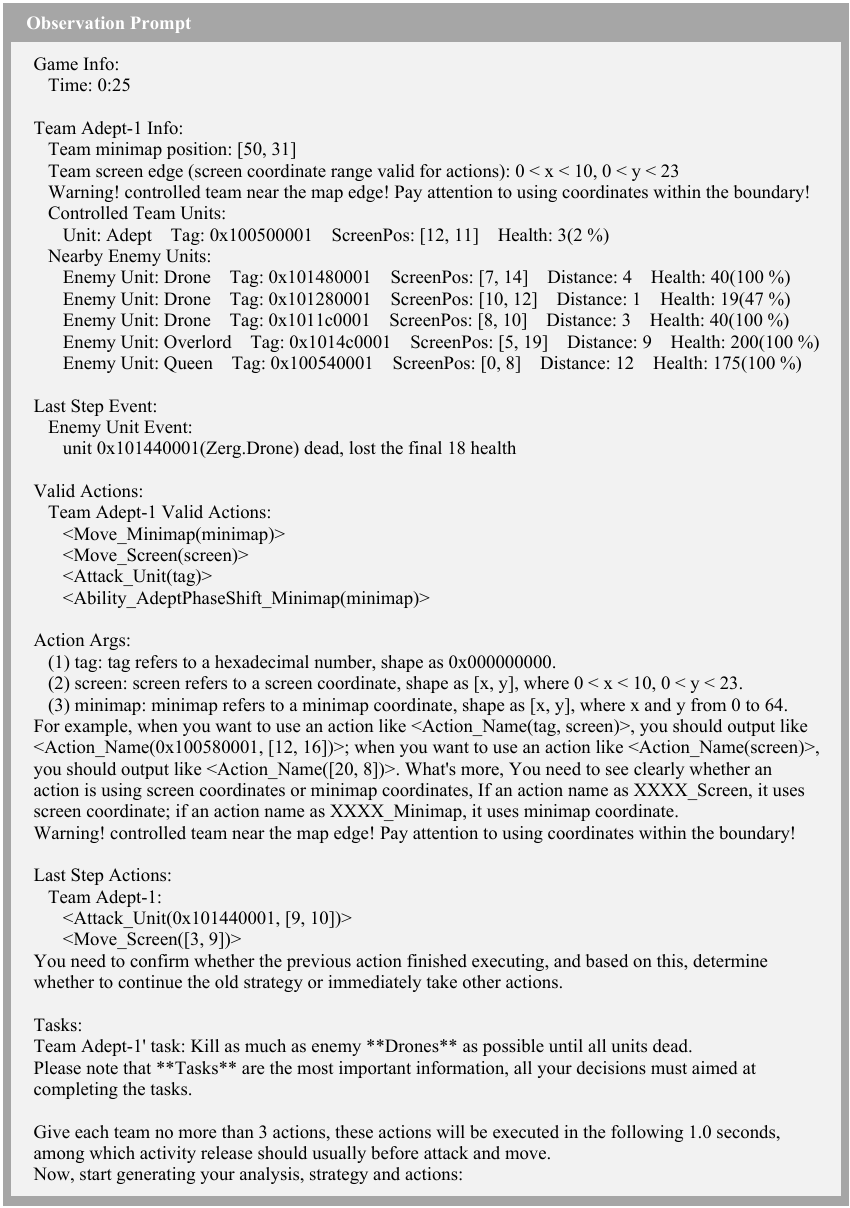}
    \caption{\textbf{An example of observation prompt.}}
  \label{fig-prompt-obs}
\end{figure}

\clearpage
\subsection*{B.2 Prompt for Action Module}

% \vspace{0.2cm}
\begin{figure}[ht]
  \centering
  \includegraphics[width=0.9\textwidth]{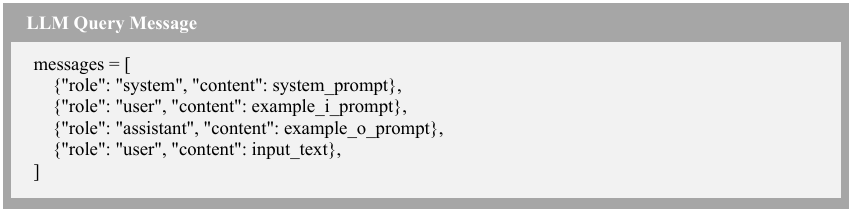}
    \caption{\textbf{System prompt for decision making.}}
  \label{fig-message}
\end{figure}

\textbf{B.2.1 Decision-making Prompt}

% \vspace{0.2cm}
\begin{figure}[ht]
  \centering
  \includegraphics[width=0.9\textwidth]{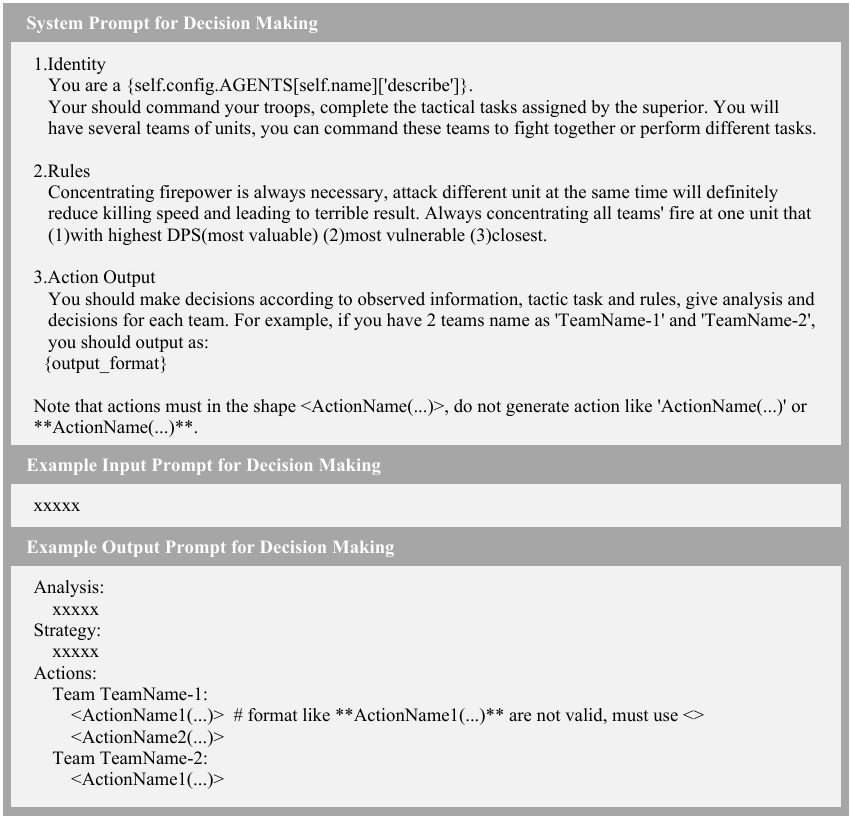}
    \caption{\textbf{System prompt for decision making.}}
  \label{fig-prompt-a}
\end{figure}

% \clearpage
% \textbf{B.2.2 Rethinking Prompt}

% \vspace{-0.2cm}
% \begin{figure}[ht]
%   \centering
%   \includegraphics[width=0.9\textwidth]{fig-prompt-rethink.pdf}
%     \caption{\textbf{System prompt for rethinking.}}
%   \label{fig-prompt-obs}
% \end{figure}

\clearpage
\subsection*{B.3 Prompt for Learning Module}

\textbf{B.3.1 Prompt for generating hypothesis}

% \vspace{-0.2cm}
\begin{figure}[ht]
  \centering
  \includegraphics[width=0.9\textwidth]{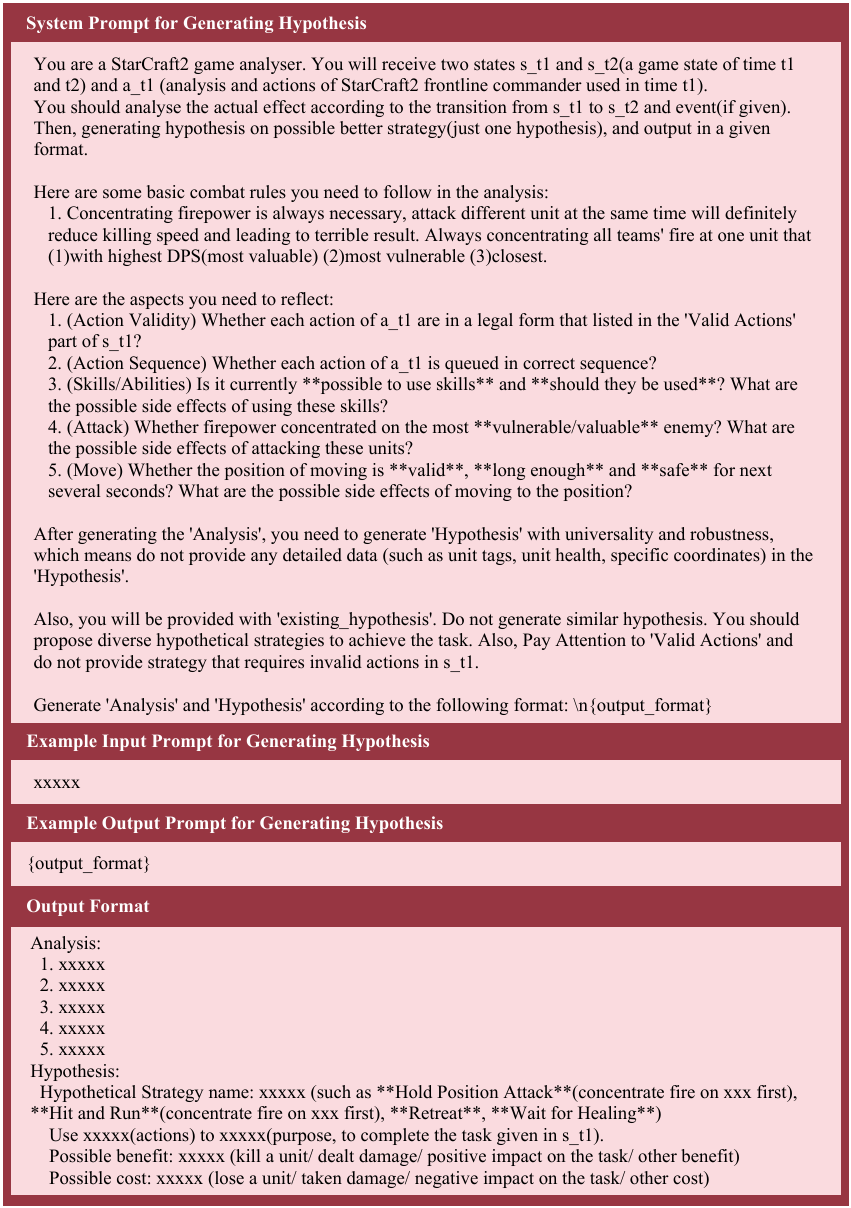}
    \caption{\textbf{System prompt, example input prompt and example output prompt for generating hypothesis.} The model is asked to generate a possible better policy, and predict its benefit and cost.}
  \label{fig-p-h-full}
\end{figure}

\clearpage
\textbf{B.3.2 Prompt for generating validation}

% \vspace{-0.2cm}
\begin{figure}[ht]
  \centering
  \includegraphics[width=0.9\textwidth]{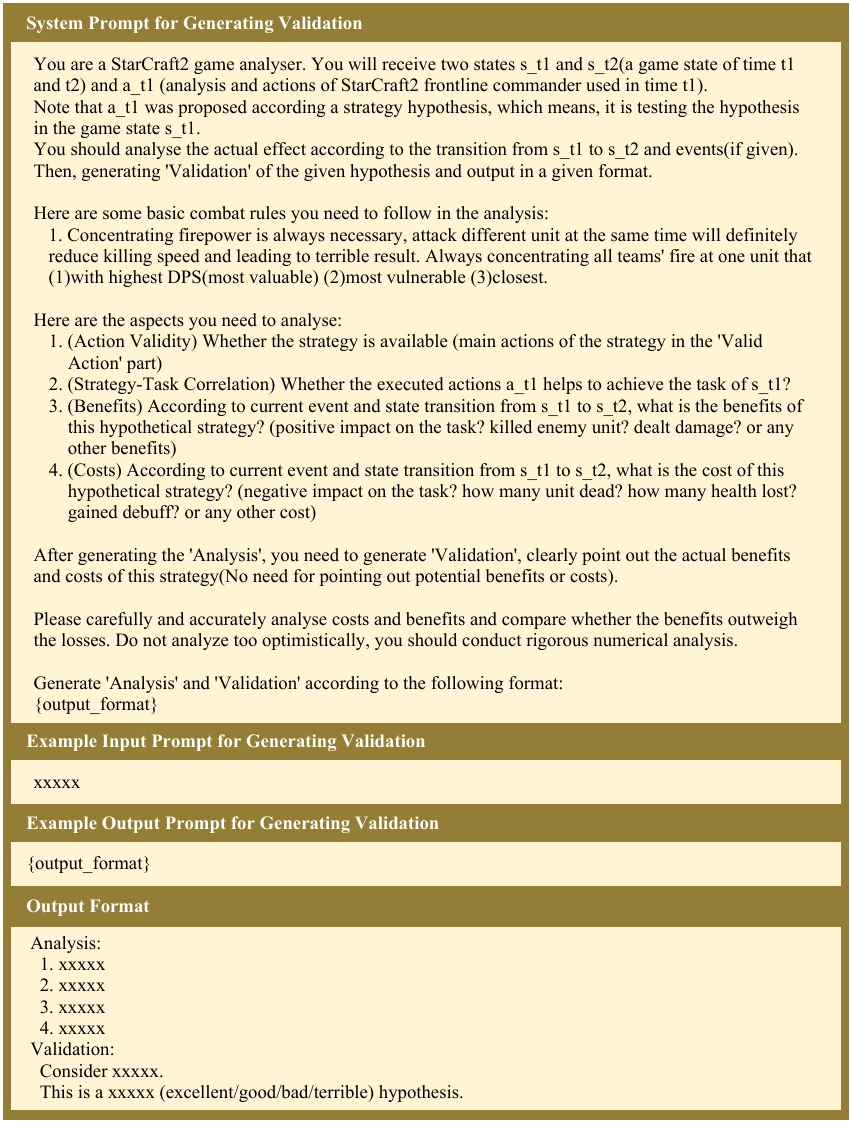}
    \caption{\textbf{System prompt, example input prompt and example output prompt for generating validations.} The model is asked to validate an existing policy, check the actual benefit and cost, analyze whether the hypothetical policy makes it closer to the task goal.}
  \label{fig-p-v-full}
\end{figure}

\clearpage
\textbf{B.3.3 Prompt for generating experience}

% \vspace{-0.2cm}
\begin{figure}[ht]
  \centering
  \includegraphics[width=0.9\textwidth]{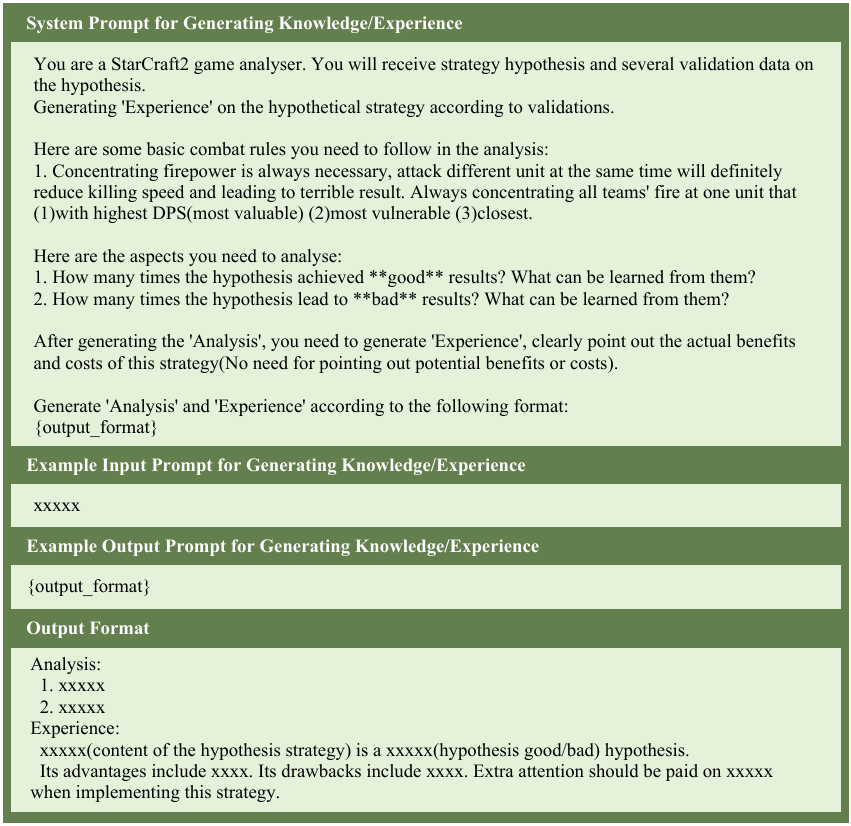}
    \caption{\textbf{System prompt, example input prompt and example output prompt for generating experience.} The model is asked to summarize experience using the hypothesis and the validation data, evaluate the performance of the policy, potential risks and benefits.}
  \label{fig-p-e-full}
\end{figure}

\clearpage
\subsection*{B.4 Example Input and Output of decision-making process}
\textbf{B.4.1 Direct Decision Making}

\begin{figure}[ht]
  \centering
  \includegraphics[width=0.92\textwidth]{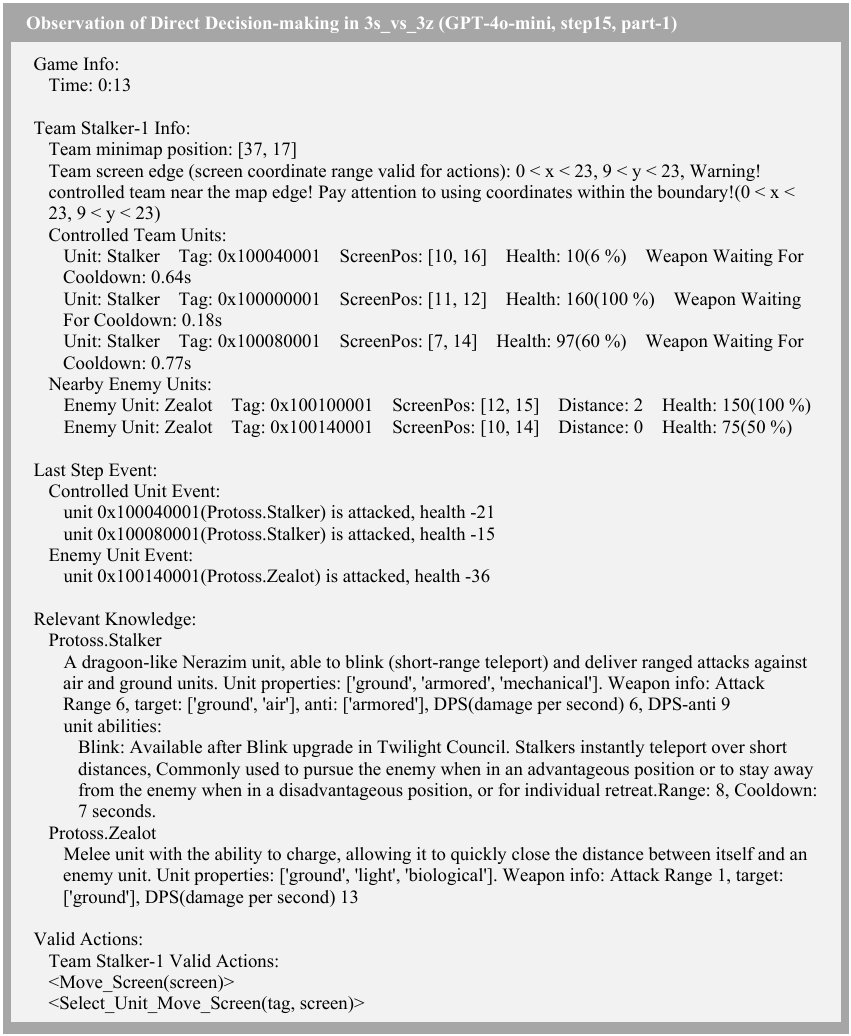}
    \caption{\textbf{Observation of direct decision-making (part1).}}
  \label{fig-ral-example1-obs1}
\end{figure}

\begin{figure}[ht]
  \centering
  \includegraphics[width=0.92\textwidth]{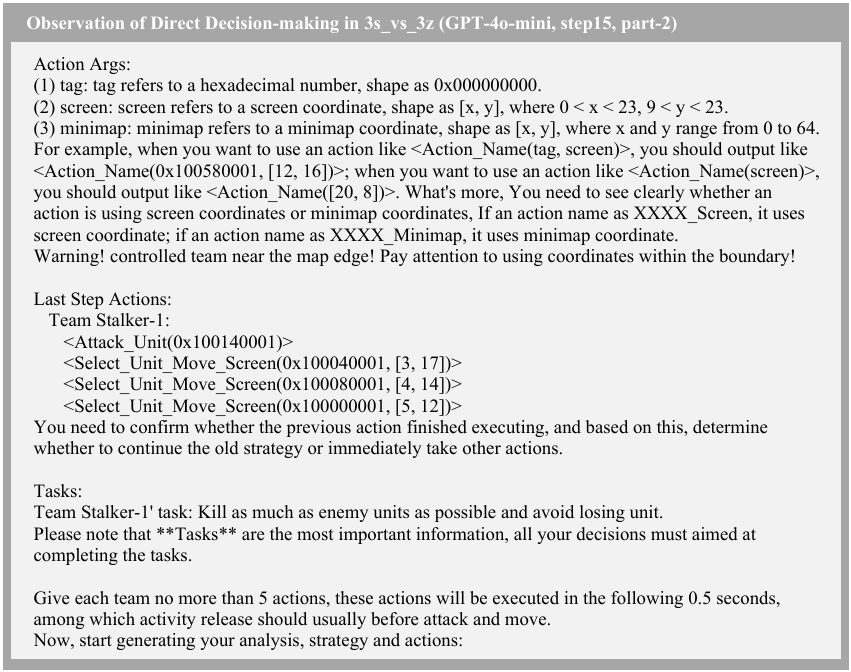}
    \caption{\textbf{Observation of direct decision-making (part2).}}
  \label{fig-ral-example1-obs2}
\end{figure}

\begin{figure}[ht]
  \centering
  \includegraphics[width=0.92\textwidth]{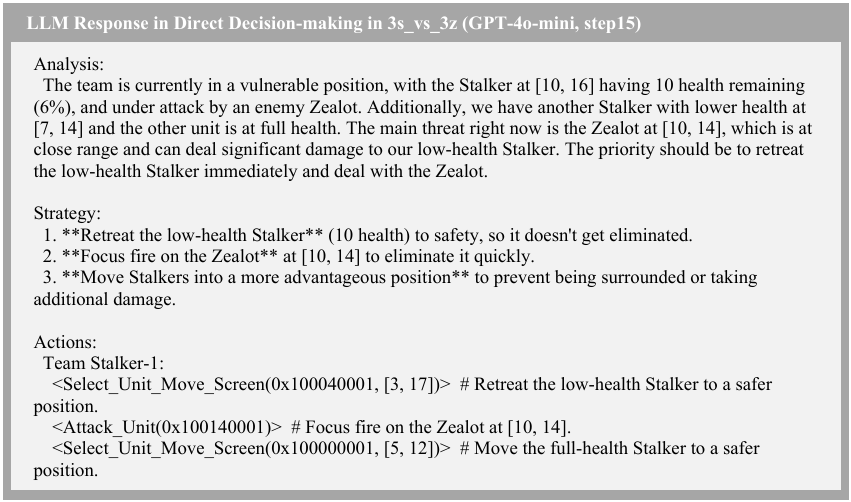}
    \caption{\textbf{Response of LLM in direct decision-making.}}
  \label{fig-ral-example1-res1}
\end{figure}

\clearpage
\textbf{B.4.2 Decision Making Follow a Hypothesis}

\begin{figure}[ht]
  \centering
  \includegraphics[width=0.92\textwidth]{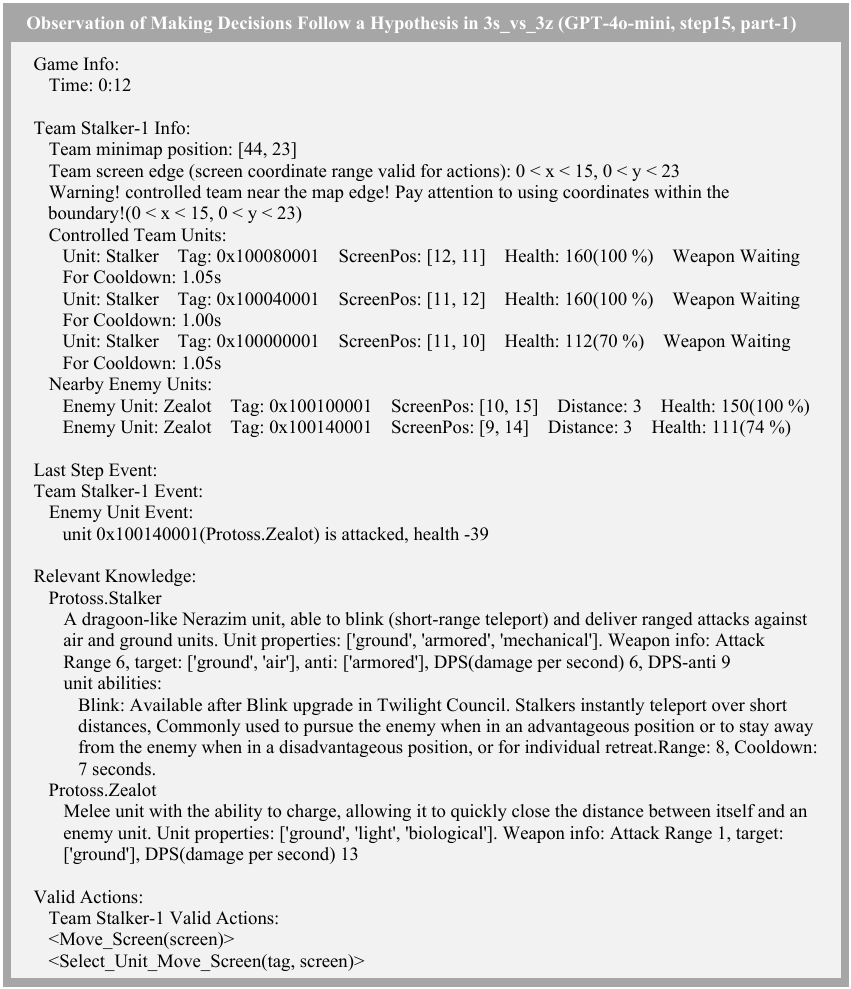}
    \caption{\textbf{Observation of decision-making that follows a hypothesis (part1).}}
  \label{fig-ral-example2-obs1}
\end{figure}

\begin{figure}[ht]
  \centering
  \includegraphics[width=0.92\textwidth]{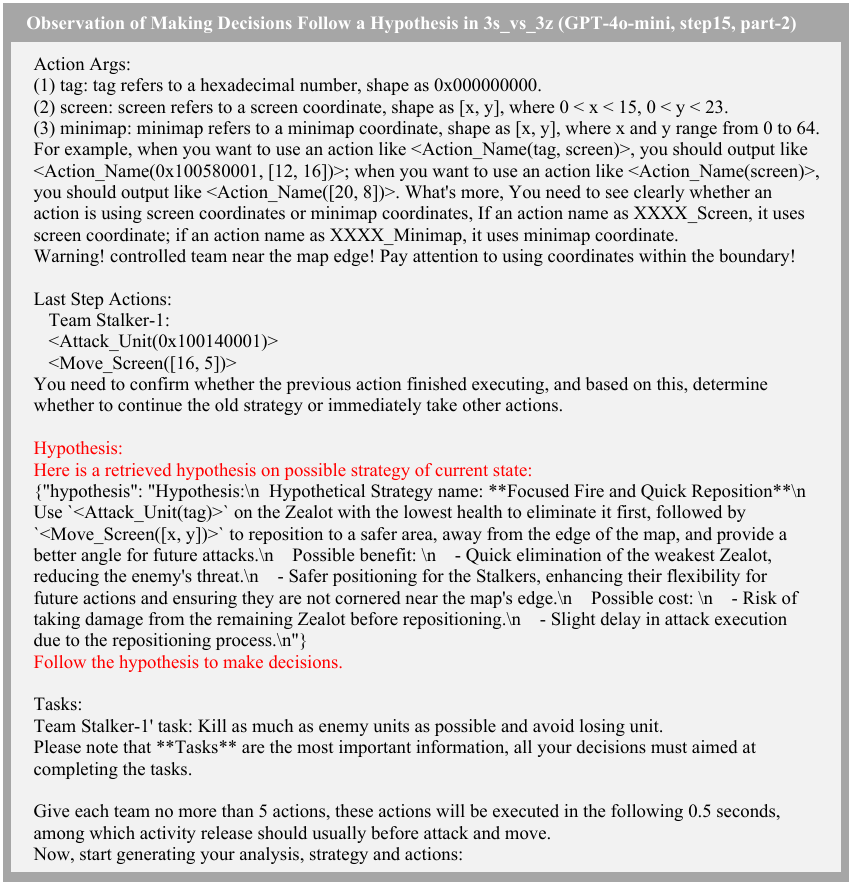}
    \caption{\textbf{Observation of decision-making that follows a hypothesis (part2).}}
  \label{fig-ral-example2-obs2}
\end{figure}

\begin{figure}[ht]
  \centering
  \includegraphics[width=0.92\textwidth]{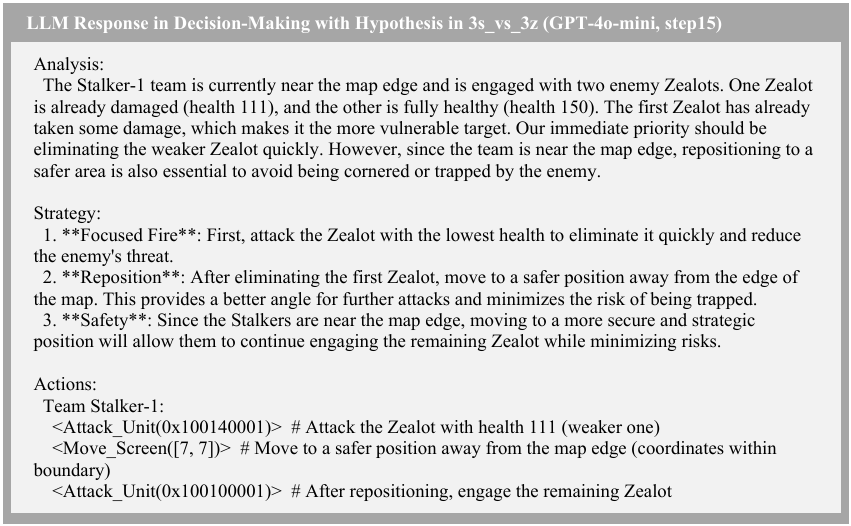}
    \caption{\textbf{Response of LLM in decision-making that follows a hypothesis.}}
  \label{fig-ral-example2-res1}
\end{figure}

\clearpage
\textbf{B.4.3 Decision Making Follow Experiences}

\begin{figure}[ht]
  \centering
  \includegraphics[width=0.92\textwidth]{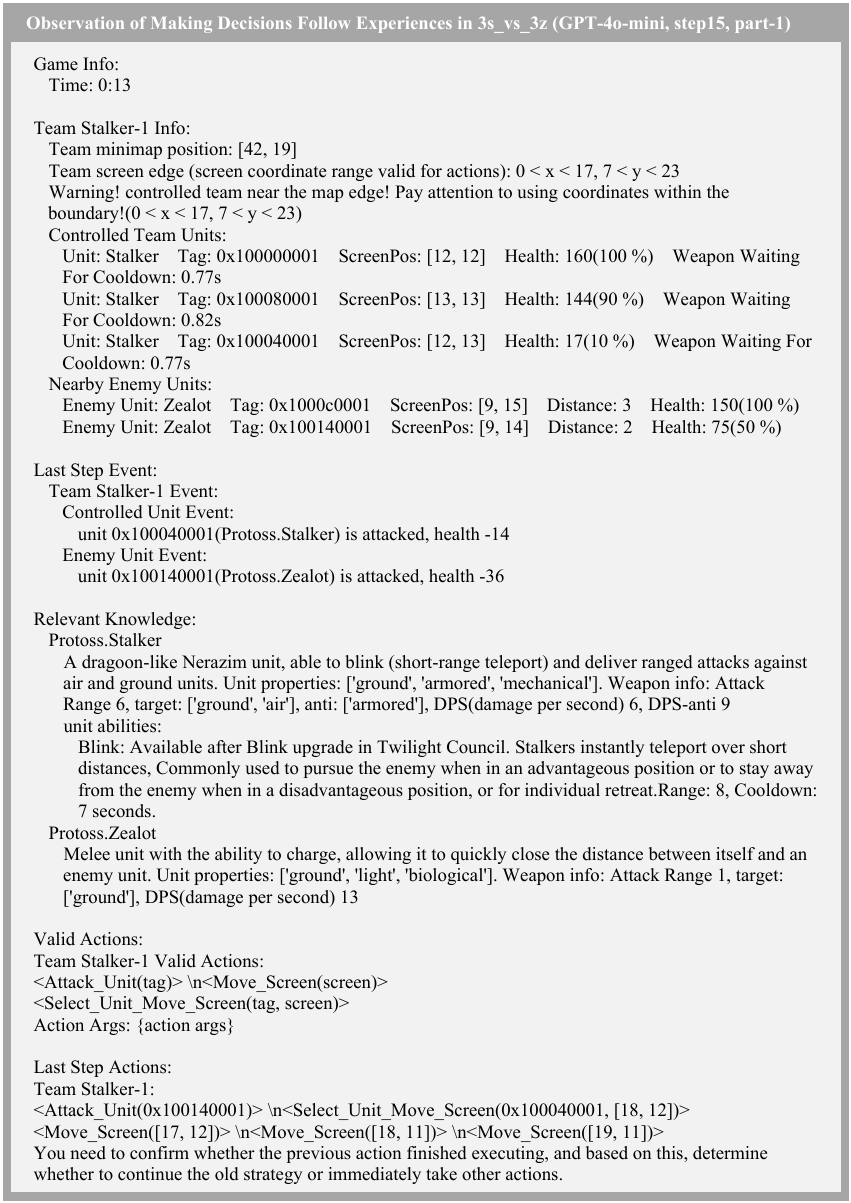}
    \caption{\textbf{Observation of decision-making that follows experiences (part1).}}
  \label{fig-ral-example3-obs1}
\end{figure}

\begin{figure}[ht]
  \centering
  \includegraphics[width=0.92\textwidth]{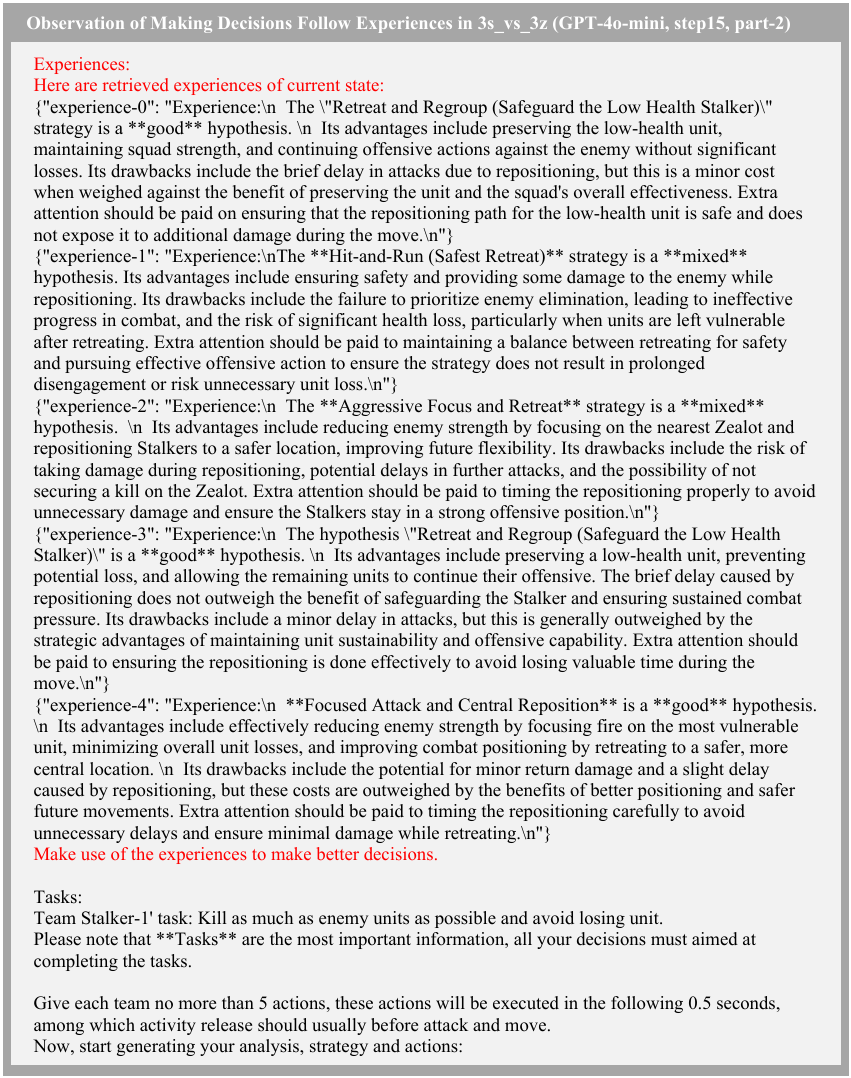}
    \caption{\textbf{Observation of decision-making that follows experiences (part2).}}
  \label{fig-ral-example3-obs2}
\end{figure}

\begin{figure}[ht]
  \centering
  \includegraphics[width=0.92\textwidth]{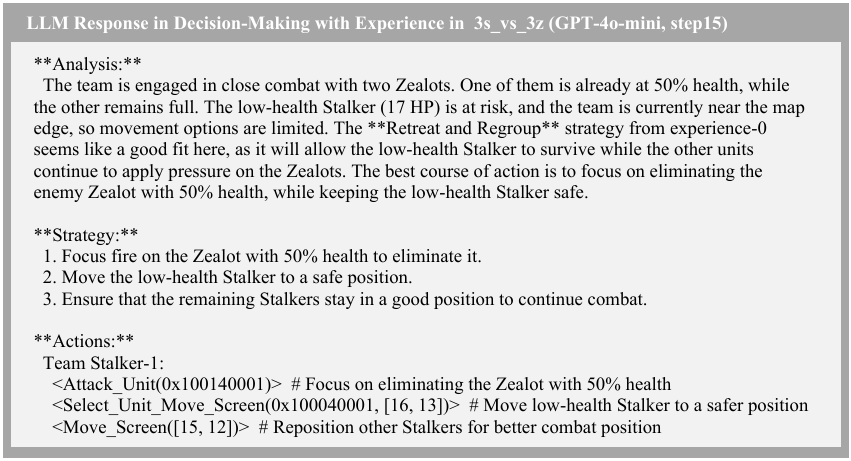}
    \caption{\textbf{Response of LLM in decision-making that follows experiences.}}
  \label{fig-ral-example3-res1}
\end{figure}

\clearpage
\subsection*{B.5 Example input and output of the learning process}

\textbf{B.5.1 Hypothesis Generation}

\begin{figure}[ht]
  \centering
  \includegraphics[width=0.92\textwidth]{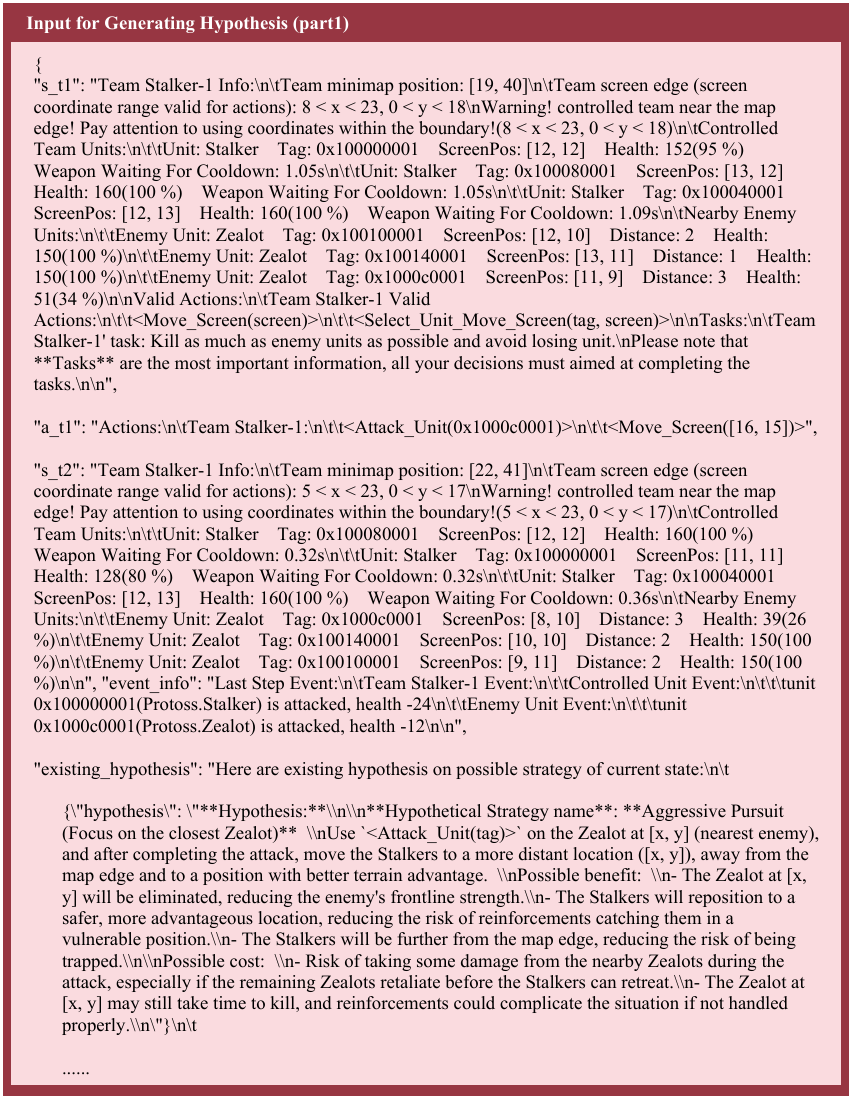}
    \caption{\textbf{Input prompt of hypothesis generation (part1).}}
  \label{fig-ral-learning-h-in1}
\end{figure}

\begin{figure}[ht]
  \centering
  \includegraphics[width=0.92\textwidth]{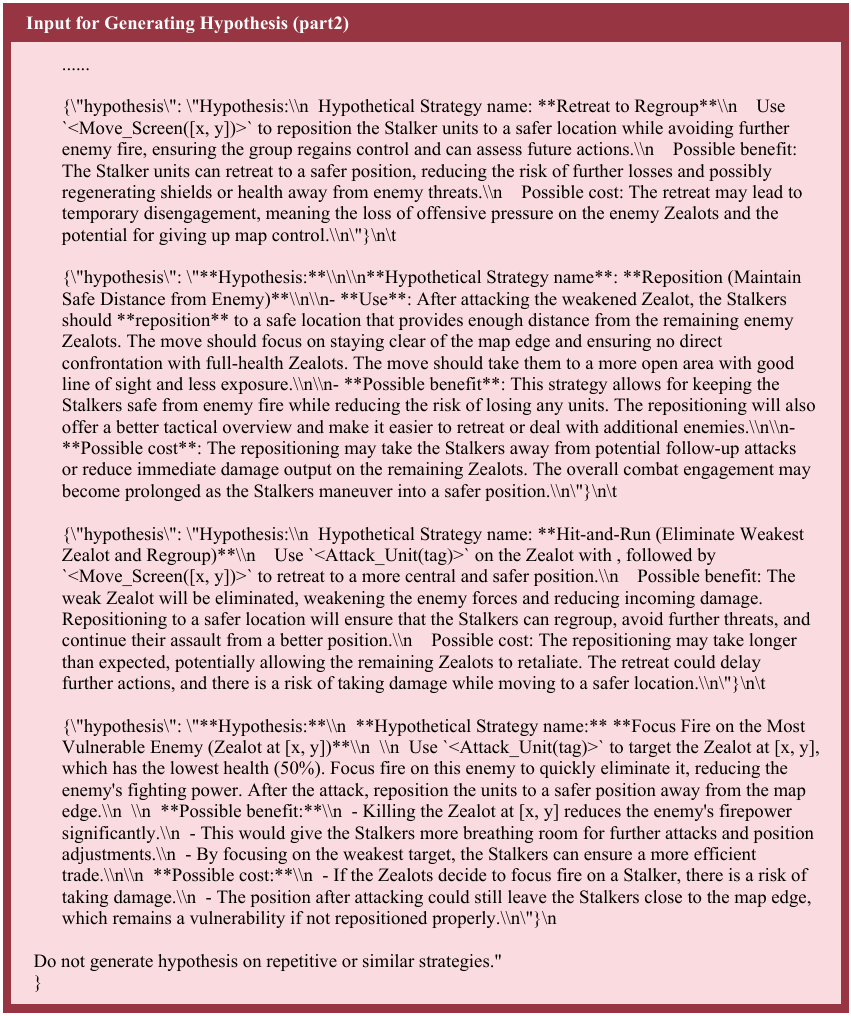}
    \caption{\textbf{Input prompt of hypothesis generation (part2).}}
  \label{fig-ral-learning-h-in2}
\end{figure}

\begin{figure}[ht]
  \centering
  \includegraphics[width=0.92\textwidth]{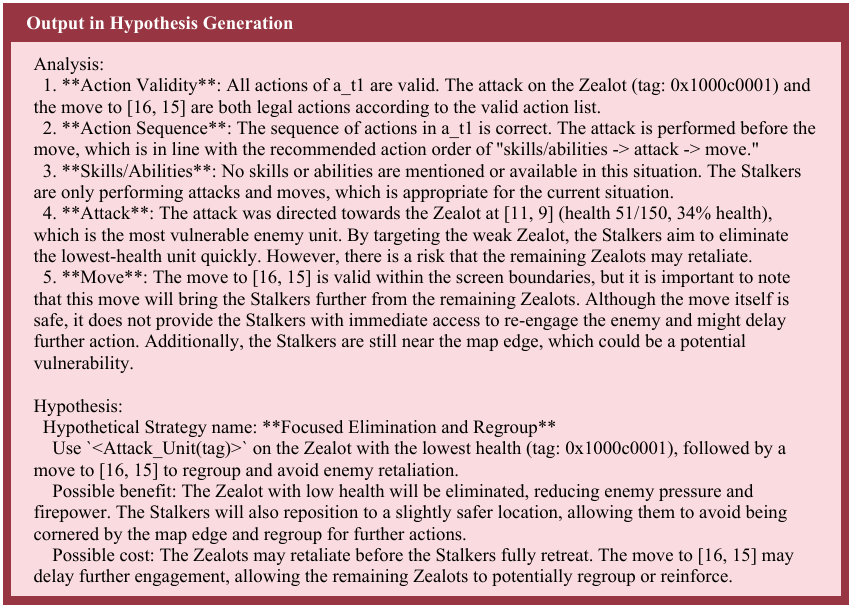}
    \caption{\textbf{Response of LLM in generating hypothesis.}}
  \label{fig-ral-learning-h-out}
\end{figure}

\begin{figure}[ht]
  \centering
  \includegraphics[width=0.92\textwidth]{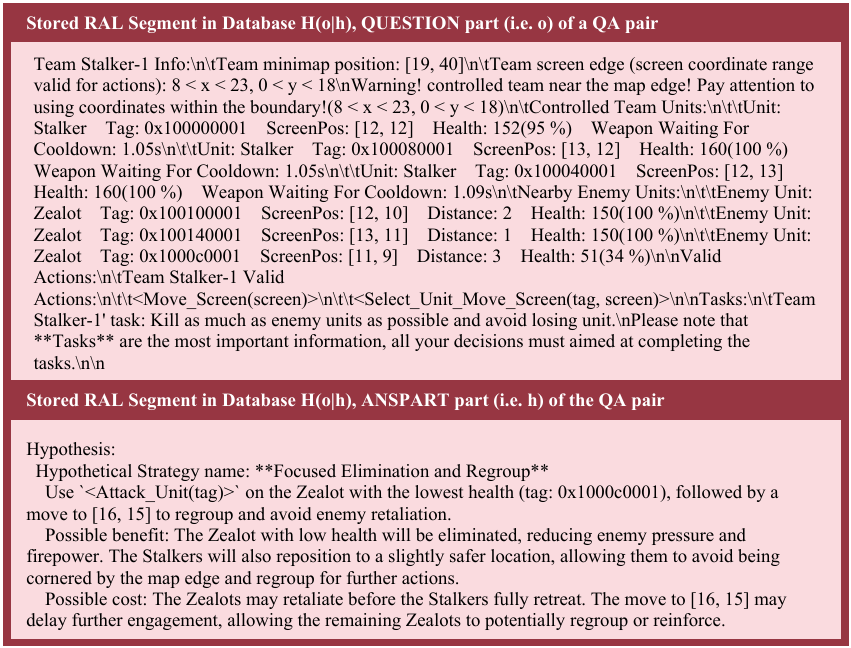}
    \caption{\textbf{Stored RAL segment in database H(o|h).}}
  \label{fig-ral-segment-h}
\end{figure}

\clearpage
\textbf{B.5.2 Validation Generation}

\begin{figure}[ht]
  \centering
  \includegraphics[width=0.92\textwidth]{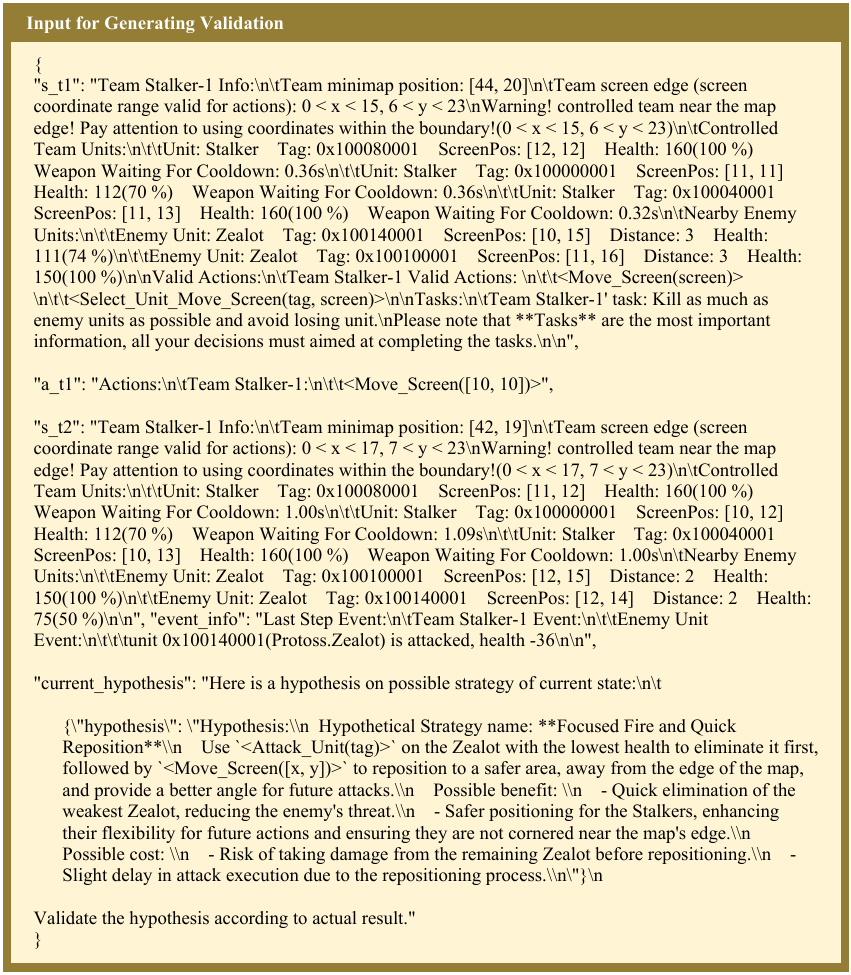}
    \caption{\textbf{Input prompt of validation generation.}}
  \label{fig-ral-learning-v-in}
\end{figure}

\begin{figure}[ht]
  \centering
  \includegraphics[width=0.92\textwidth]{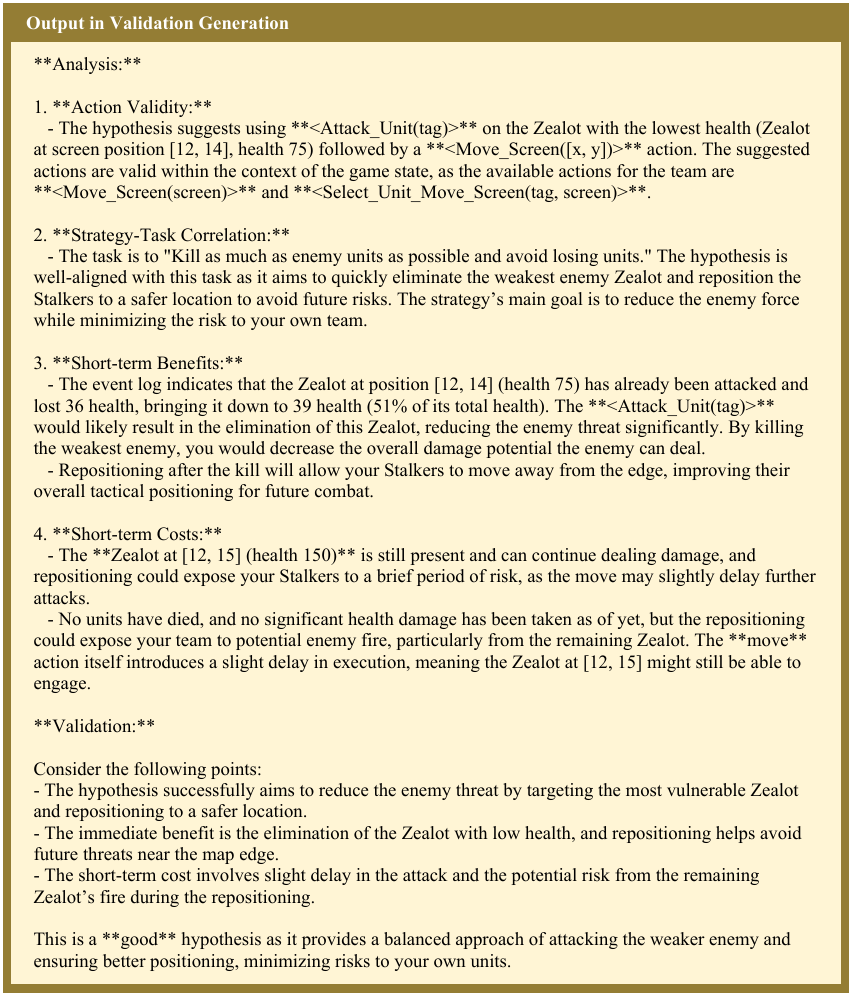}
    \caption{\textbf{Response of LLM in validation hypothesis.}}
  \label{fig-ral-learning-v-out}
\end{figure}

\begin{figure}[ht]
  \centering
  \includegraphics[width=0.92\textwidth]{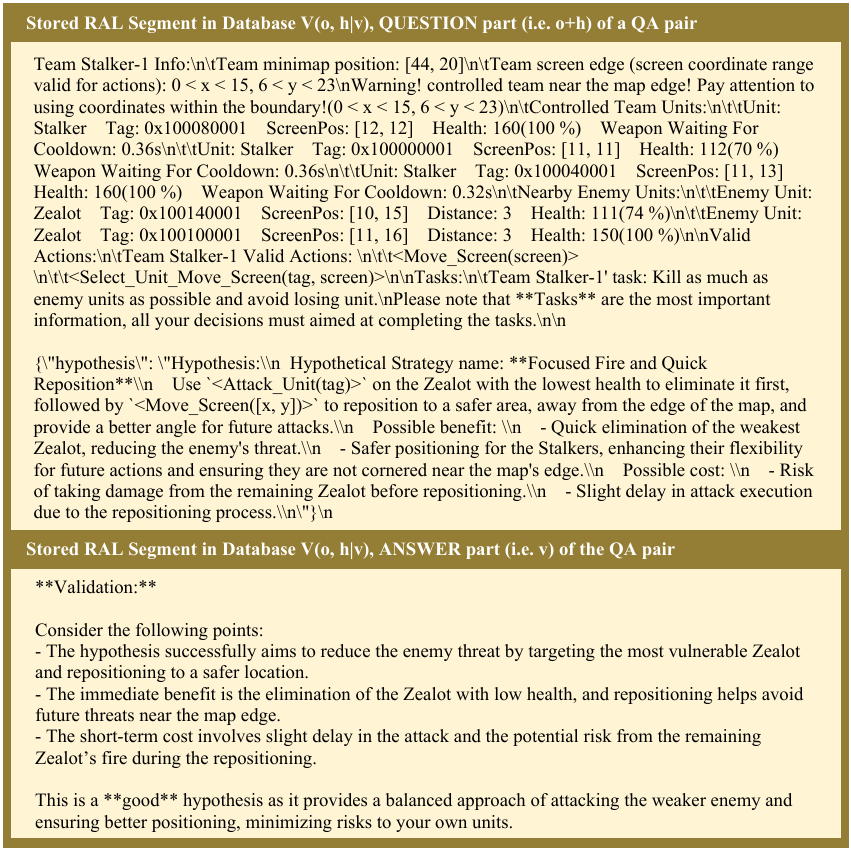}
    \caption{\textbf{Stored RAL segment in database V(o,h|v).}}
  \label{fig-ral-segment-v}
\end{figure}

\clearpage
\textbf{B.5.3 Experience Generation}

\begin{figure}[ht]
  \centering
  \includegraphics[width=0.92\textwidth]{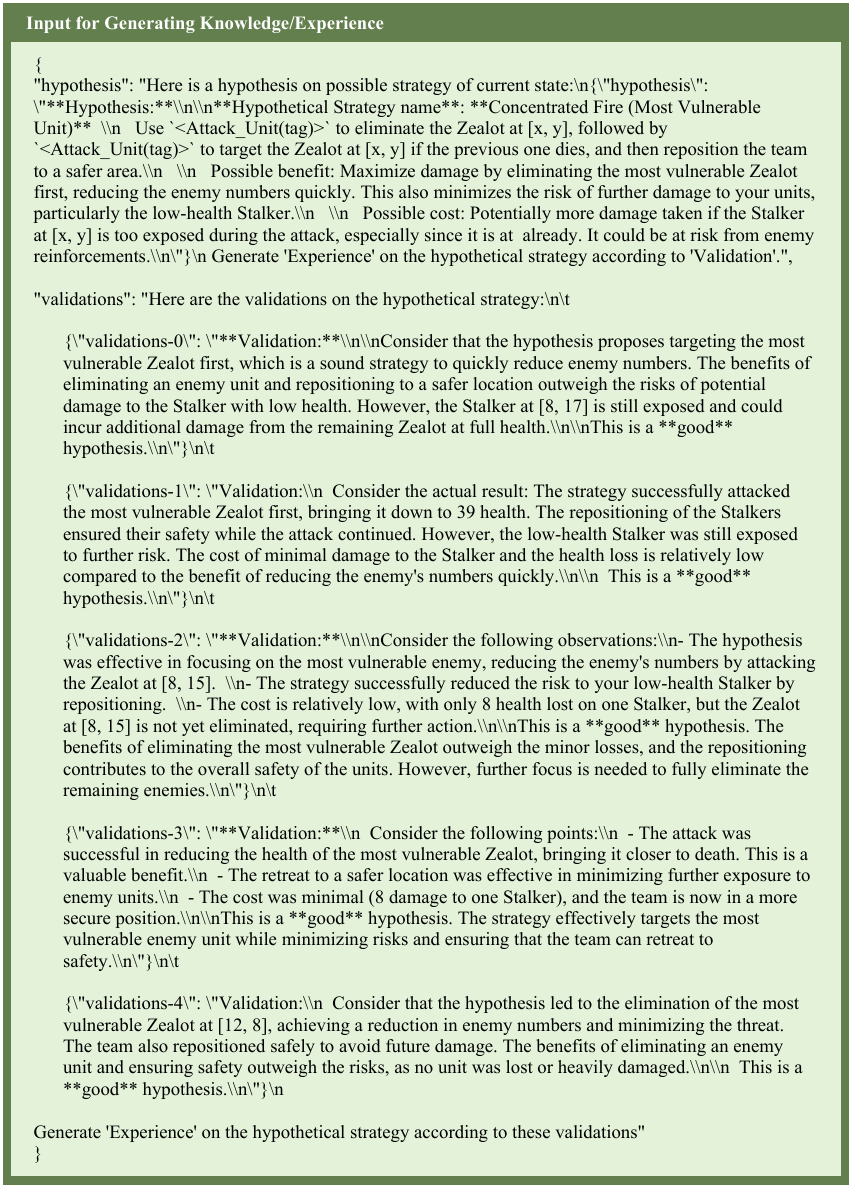}
    \caption{\textbf{Input prompt of experience generation.}}
  \label{fig-ral-learning-e-in}
\end{figure}

\begin{figure}[ht]
  \centering
  \includegraphics[width=0.92\textwidth]{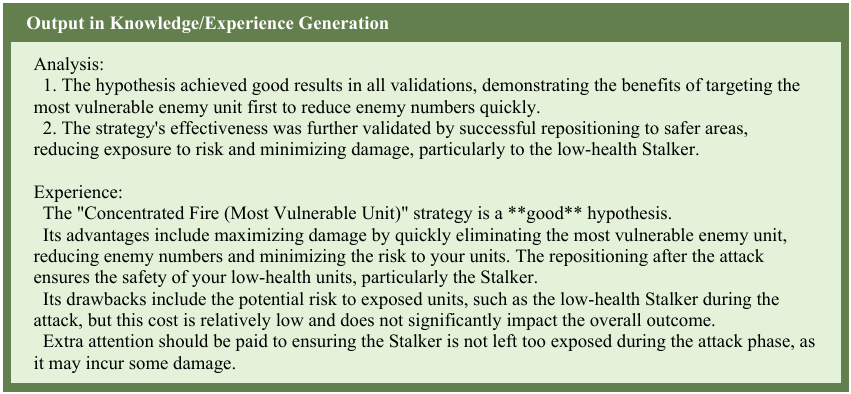}
    \caption{\textbf{Response of LLM in experience hypothesis.}}
  \label{fig-ral-learning-e-out}
\end{figure}

\begin{figure}[ht]
  \centering
  \includegraphics[width=0.92\textwidth]{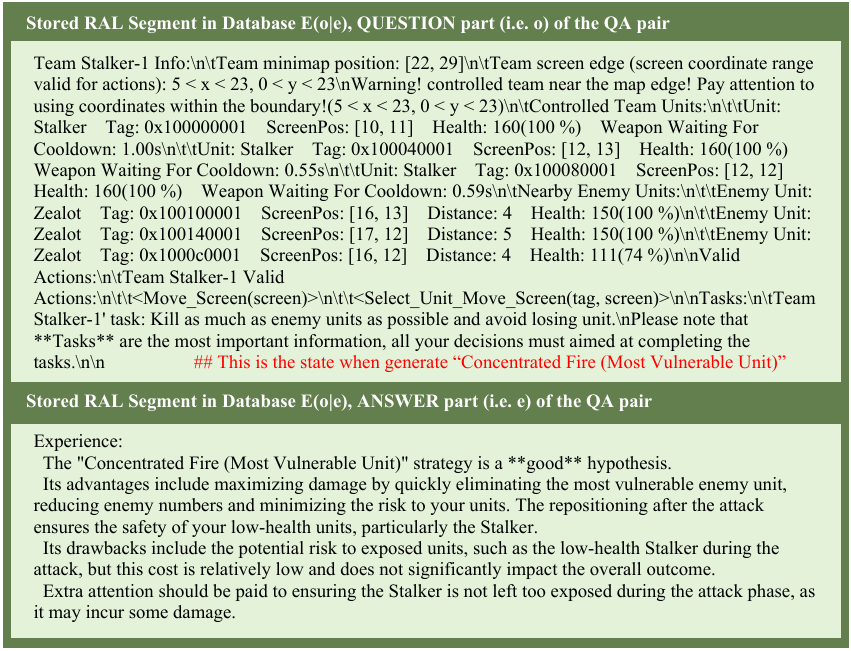}
    \caption{\textbf{Stored RAL segment in database E(o|e).}}
  \label{fig-ral-segment-e}
\end{figure}

\clearpage
\section*{Appendix C. Experiment Settings}
\setcounter{figure}{0}
\setcounter{table}{0}
\renewcommand{\thefigure}{C\arabic{figure}}
\renewcommand{\thetable}{C\arabic{table}}

\subsection*{C.1 Settings of experimental tasks}

\begin{table}[ht]
\vspace{0.2cm}
\caption{System settings}\label{tabc1}
\begin{center}
\vspace{-0.2cm}
\small
\renewcommand\arraystretch{1.2}
\begin{tabular}{p{3cm} p{6.3cm}  p{4.3cm}}
% \hline
% & \multicolumn{6}{c}{Task name} \\
\toprule
Module & Details    & Minimum requirements\\
\midrule
System  & Windows-11 & Windows-10\\
CPU  & i9-14900k, 24 cores 32 threads & 8 core\\
GPU  & GeForce RTX 4090, 24G & GeForce GTX 1080\\
RAM  & 64G & 8G\\
SSD  & 2T  & 100G\\
Starcraft II & Version 9.0.14(93333) & Version 9.0.14(93333)\\
% Python  & Conda, python==3.9.13 & --\\
LLM-PySC2  & A developed version based on LLM-PySC2 v0.1, with some changes in actions and observations. & -- \\
Maps & Mostly provided by LLM-PySC2 v0.1, except the 4s\_blink\_5r/1R4r maps and the 3ph\_harass map with ProtossAirAttackLevel1 research added . & --\\

\bottomrule
\end{tabular}
\end{center}
\end{table}

\begin{table}[ht]
\vspace{0.2cm}
\caption{Unit settings}\label{tabc2}
\begin{center}
\vspace{-0.2cm}
\small
\renewcommand\arraystretch{1.2}
\begin{tabular}{p{2.5cm} p{2.8cm} p{2.8cm} p{2.3cm} p{2.3cm}}
% \hline
% & \multicolumn{6}{c}{Task name} \\
\toprule
 Task name &  Controlled  & Enemy  & Life C/E  & DPS C/E \\
\midrule
% 2s3z                      & 2 Stalkers, 3 Zealots & 2 Stalkers, 3 Zealots & 770/770=1.00 & 50.8/50.8=1.00 \\
% 2s\_vs\_1sc               & 2 Stalkers & 1 SpineCrawler & 320/300=1.07 &  13.9/22.7=0.61 \\
3s\_vs\_3z                & 3 Stalkers & 3 Zealots & 480/450=1.07 & 13.9/36.9=0.38 \\ 
3s\_vs\_4z                & 3 Stalkers & 4 Zealots & 480/600=0.80 & 13.9/49.2=0.28 \\
3s\_vs\_5z                & 3 Stalkers & 5 Zealots & 480/750=0.64 & 13.9/61.5=0.23 \\
2a\_harass                & 2 Adapts & 2 Queens + 12 Drones & - & - \\
3ph\_harass               & 3 Phoenixes & 2 Queens + 12 Drones & - & - \\
4s\_blink\_vs\_5r         & 4 Stalkers (Blink On) & 5 Roaches & 640/725=0.88 & 38.5/55.9=0.69 \\
4s\_blink\_vs\_1R4r       & 4 Stalkers (Blink On) & 1 Ravager, 4 Roaches & 640/700=0.91 & 38.5/54.8=0.70 \\
\bottomrule
\end{tabular}
\end{center}
\end{table}

\begin{table}[ht]
\vspace{0.2cm}
\caption{Victory conditions}\label{tabc3}
\begin{center}
\vspace{-0.2cm}
\small
\renewcommand\arraystretch{1.2}
\begin{tabular}{p{3cm} p{9cm} p{1.6cm}}
% \hline
% & \multicolumn{6}{c}{Task name} \\
\toprule
Task name &  victory condition & max time\\
\midrule
% 2s3z                      & defeat all enemies, before all the controlled units dead & - \\
% 2s\_vs\_1sc               & defeat all enemies, before all the controlled units dead & - \\
3s\_vs\_3z                & defeat all enemies, before all the controlled units dead & - \\ 
3s\_vs\_4z                & defeat all enemies, before all the controlled units dead & - \\
3s\_vs\_5z                & defeat all enemies, before all the controlled units dead & - \\
2a\_harass                & kill at least 7 workers, before all the controlled units dead & 1min\\
3ph\_harass               & kill at least 7 workers, before all the controlled units dead & 1min\\
4s\_blink\_vs\_5r         & defeat all 5 Roaches, before all the controlled units dead & 1min \\
4s\_blink\_vs\_1R4r       & defeat all 4 Roaches and the Ravager, before all the controlled units dead & 1min \\
\bottomrule
\end{tabular}
\end{center}
\end{table}

\begin{table}[ht]
\vspace{0.2cm}
\caption{Agent settings}\label{tabc4}
\begin{center}
\vspace{-0.2cm}
\small
\renewcommand\arraystretch{1.2}
\begin{tabular}{p{3cm} p{1.5cm} p{9.1cm}}
% \hline
% & \multicolumn{6}{c}{Task name} \\
\toprule
Task name &  num agent & controlled unit teams\\
\midrule
% 2s3z                      & 1 & Team 'Stalker-1': 2 Stalkers; Team 'Zealot-1': 3 Zealots \\
% 2s\_vs\_1sc               & 1 & Team 'Stalker-1': 1 Stalker; \ Team 'Stalker-2': 1 Stalker \\
3s\_vs\_3z                & 1 & Team 'Stalker-1': 3 Stalkers \\ 
3s\_vs\_4z                & 1 & Team 'Stalker-1': 3 Stalkers \\
3s\_vs\_5z                & 1 & Team 'Stalker-1': 3 Stalkers \\
2a\_harass                & 1 & Team 'Adept-1': 3 Stalkers; \ Team 'AdeptPhase-1': 2 AdeptShadows\\
3ph\_harass               & 1 & Team 'Phoenix-1': 3 Phoenixes\\
4s\_blink\_vs\_5r         & 1 & Team 'Stalker-1': 4 Stalkers \\
4s\_blink\_vs\_1R4r       & 1 & Team 'Stalker-1': 4 Stalkers \\
\bottomrule
\end{tabular}
\end{center}
\end{table}

\begin{table}[ht]
\vspace{0.2cm}
\caption{Action space}\label{tabc5}
\begin{center}
\vspace{-0.2cm}
\small
\renewcommand\arraystretch{1.2}
\begin{tabular}{p{3cm} p{3cm} p{7.5cm}}
% \hline
% & \multicolumn{6}{c}{Task name} \\
\toprule
Task name & Unit name & Actions \\
% \midrule
% 2s3z                      & Zealot & <Attack\_Unit(tag)> \\ 
%                           & Stalker & <Attack\_Unit(tag)> \\
%                           &  & <Move\_Screen(screen)> \\
% \midrule
% 2s\_vs\_1sc               & Stalker & <Attack\_Unit(tag)> \\
%                           &  & <Move\_Screen(screen)> \\
\midrule
3s\_vs\_3z                & Stalker & <Attack\_Unit(tag)> \\
3s\_vs\_4z                &  & <Move\_Screen(screen)> \\
3s\_vs\_5z                &  & <Select\_Unit\_Move\_Screen(tag, screen)> \\
\midrule
2a\_harass                & Adept & <Attack\_Unit(tag)> \\ 
                          &  & <Move\_Screen(screen)> \\
                          &  & <Move\_Minimap(minimap)> \\
                          &  & <Select\_Unit\_Attack\_Unit(tag, tag)> \\
                          &  & <Select\_Unit\_Move\_Screen(tag, screen)> \\
                          & AdeptPhase & <Move\_Screen(screen)> \\ 
                          &  & <Move\_Minimap(minimap)> \\
\midrule
3ph\_harass               & Phoenix & <Attack\_Unit(tag)> \\ 
                          &  & <Move\_Screen(screen)> \\ 
                          &  & <Move\_Minimap(minimap)> \\
                          &  & <Select\_Phoenix\_Ability\_GravitonBeam\_Unit(tag, tag)> \\
                          &  & <Cancel\_GravitonBeam\_For\_Phoenix(tag, screen)> \\
\midrule
4s\_blink\_vs\_5r         & Stalker & <Attack\_Unit(tag)> \\ 
4s\_blink\_vs\_1R4r       &  & <Move\_Screen(screen)> \\ 
                          &  & <Ability\_Blink\_Screen(screen)> \\
                          &  & <Select\_Unit\_Attack\_Unit(tag, tag)> \\
                          &  & <Select\_Unit\_Move\_Screen(tag, screen)> \\
                          &  & <Select\_Unit\_Blink\_Screen(tag, screen)> \\
\bottomrule
\end{tabular}
\end{center}
\end{table}

\clearpage
\subsection*{C.2 RAG settings}

\begin{table}[ht]
\vspace{0.2cm}
\caption{RAG settings}\label{tabc6}
\begin{center}
\vspace{-0.2cm}
\small
\renewcommand\arraystretch{1.2}
\begin{tabular}{p{3cm} p{10.6cm}}
% \hline
% & \multicolumn{6}{c}{Task name} \\
\toprule
Module & Details \\
\midrule
RAG framework  & Dify, a modified version based on v0.11.2, with the limitation on segment length and search length removed. The first limitation in dify compose file line 249 INDEXING\_MAX\_SEGMENTATION\_TOKENS\_LENGTH. The second limitation in dify source code dify.api.services.hit\_testing\_service.py line 165. All the Dify services are provided by docker-desktop of version 4.35.1 (173168), initialize the container through the compose file\\
RAG embedding  & GLM Embedding-3\\
RAG retrieve mothod   & Hybrid search, weight 0.5, reranking disabled \\
RAG databases   & 4 databases in total, query-answer mode, collect data for each scenario into independent documents, store query-answer pair in a segment of a document\\

\bottomrule
\end{tabular}
\end{center}
\end{table}

\subsection*{C.3 LLM-SSCL settings}

\begin{table}[ht]
\vspace{0.2cm}
\caption{Database Hyper-parameters}\label{tabc7}
\begin{center}
\vspace{-0.2cm}
\small
\renewcommand\arraystretch{1.2}
\begin{tabular}{p{3cm} p{3cm} p{1.5cm} p{1.5cm} p{3cm}}

\toprule
Module name & Task Group & Threshold & Top-k & update rate\\
\midrule
$H(o|h)$    & LLM-SMAC tasks   & 0.995  & 5  &  0 \\
            & LLM-PySC2 tasks  & 0.99   & 5  &  0 \\
$V(o,h|v)$  & LLM-SMAC tasks   & 0.97   & 5  &  0.1 \\
            & LLM-PySC2 tasks  & 0.97   & 5  &  0.1 \\
$E(o|e)$    & LLM-SMAC tasks   & 0.995  & 5  &  0.1 \\
            & LLM-PySC2 tasks  & 0.99   & 5  &  0.1 \\
% $E2(o,h|e)$ & LLM-SMAC tasks   & 0.97   & 1  &  whenever E1 update \\
%             & LLM-PySC2 tasks  & 0.97   & 1  &  whenever E1 update \\

\bottomrule
\end{tabular}
\end{center}
\end{table}

% \begin{table}[ht]
% \vspace{0.2cm}
% \caption{Decision-Making Prompt}\label{tabc8}
% \begin{center}
% \vspace{-0.2cm}
% \small
% \renewcommand\arraystretch{1.2}
% \begin{tabular}{p{3cm} p{6cm}}

% \toprule
% Task name & Decision Prompt\\
% \midrule
% % 2s3z                      & basic decision prompt with rethinking prompt  \\
% % 2s\_vs\_1sc               & basic decision prompt with rethinking prompt  \\
% 3s\_vs\_3z                & basic decision prompt with rethinking prompt  \\
% 3s\_vs\_4z                & basic decision prompt with rethinking prompt  \\
% 3s\_vs\_5z                & basic decision prompt with rethinking prompt  \\
% 2a\_harass                & basic decision prompt \\
% 3ph\_harass               & basic decision prompt \\
% 4s\_blink\_vs\_5r         & basic decision prompt \\
% 4s\_blink\_vs\_1R4r       & basic decision prompt \\
% \bottomrule
% \end{tabular}
% \end{center}
% \end{table}

% \clearpage
% \section*{Appendix D. Detailed Experimental Results}
% \setcounter{figure}{0}
% \renewcommand{\thefigure}{C\arabic{figure}}

% \clearpage
% \section*{Appendix E. LLM decision details analysis}
% \setcounter{figure}{0}
% \renewcommand{\thefigure}{C\arabic{figure}}

\end{document}